\RequirePackage{fix-cm}
\documentclass[twocolumn]{svjour3}          
\smartqed  
\usepackage{graphicx}
\usepackage[authoryear]{natbib}
\usepackage{amsmath}
\usepackage{array}
\usepackage{subfigure}
\usepackage{fixltx2e}
\usepackage{url}
\usepackage{caption}
\usepackage{tabularx}
\usepackage{times} 
\usepackage{amsmath} 
\usepackage{amssymb}  
\usepackage{siunitx}

\usepackage{multirow,color}

\usepackage{url}
\usepackage{hyperref}
\usepackage{microtype}
\usepackage{gensymb}
\usepackage{enumerate}

\begin{document}

\title{Task-centric Optimization of Configurations for Assistive Robots
\thanks{We thank Henry and Jane Evans for providing input on assistive tasks and actuated beds. We also thank Daehyung Park and Zackory Erickson for their valuable feedback. This work was supported in part by the National Institute on Disability, Independent Living, and Rehabilitation Research (NIDILRR), grant 90RE5016-01-00 via RERC TechSAge,and by NSF Awards IIS-1514258 and IIS-1150157. Dr. Kemp is a cofounder, a board member, an equity holder, and the CTO of Hello Robot, Inc., which is developing products related to this research. This research could affect his personal financial status. The terms of this arrangement have been reviewed and approved by Georgia Tech in accordance with its conflict of interest policies.}
}


\author{Ariel Kapusta         \and
        Charles C. Kemp 
}


\institute{A. Kapusta \at
              Georgia Institute of Technology
              \email{akapusta@gatech.edu}           
           \and
           C.C. Kemp \at
              Georgia Institute of Technology
}

\date{Received: date / Accepted: date}

\maketitle

\begin{abstract}
Robots can provide assistance to a human by moving objects to locations around the person's body. With a well chosen initial configuration, a robot can better reach locations important to an assistive task despite model error, pose uncertainty and other sources of variation. However, finding effective configurations can be challenging due to complex geometry, a large number of degrees of freedom, task complexity and other factors. 
We present task-centric optimization of robot configurations (TOC), which is an algorithm that finds configurations from which the robot can better reach task-relevant locations and handle task variation. Notably, TOC can return more than one configuration that when used sequentially enable a simulated assistive robot to reach more task-relevant locations. TOC performs substantial offline computation to generate a function that can be applied rapidly online to select robot configurations based on current observations. 
TOC explicitly models the task, environment, and user, and implicitly handles error using representations of robot dexterity. 
We evaluated TOC in simulation with a PR2 assisting a user with 9 assistive tasks in both a wheelchair and a robotic bed. 
TOC had an overall average success rate of 90.6\% compared to 50.4\%, 43.5\%, and 58.9\% for three baseline methods from literature.
We additionally demonstrate how TOC can find configurations for more than one robot and
can be used to assist in designing or optimizing environments.

\end{abstract}

\begin{figure}[t!]
\centering
\includegraphics[width=\columnwidth]{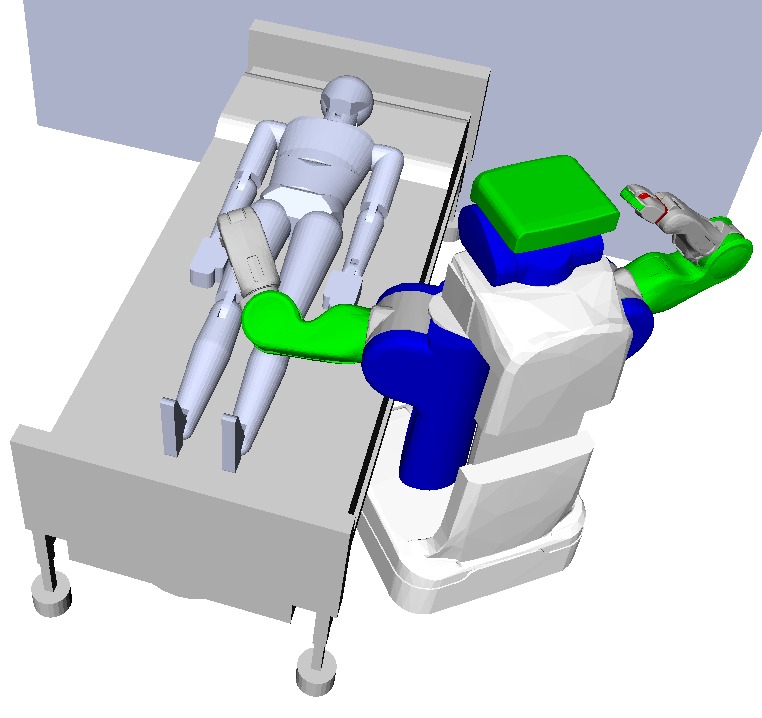}
\caption{TOC can select a configuration for the PR2 and the robotic bed so the PR2
can better reach task-relevant locations. 
This figure shows the configuration for the task of a PR2 cleaning the legs of a person in a robotic bed.
}
\label{fig:bathing_legs_autobed}
\end{figure}

\section{Introduction}\label{sec:introduction}
Robotic assistance with activities of daily living (ADLs) \citep{adls_1990}
could potentially enable people to be more independent. This may improve quality of life \citep{vest2011disability, andersen2004ability} and help address societal challenges, such as aging populations, high health care costs, and shortages of health care workers found in the United States and other countries \citep{iom_2008,Goodin2003}. 

Many specialized assistive devices can help people with motor impairments perform ADLs on their own \citep{device_website_1,device_website_2}. Specialized robots, such as desktop feeding devices, have been successful for a narrow range of assistive tasks when placed in fixed and designated positions with respect to the user \citep{topping1998dhr,handy1}. 
The choice of where to place such robots is important, as it can impact the robot's ability to provide effective assistance. 
General-purpose mobile manipulators have the advantage of mobility, which may allow them to provide assistance across a wide range of tasks, users, and environments \citep{chen2013robots}. However, this mobility introduces additional complexity. 
For a mobile manipulator to provide assistance to a person, it may first need to get close and be able to reach task-relevant locations. For mobile manipulators, the problem of selecting the robot's base pose may arise repeatedly as 
it moves around performing tasks. 
In this paper we use the more general term, configuration, which includes the robot's base pose as well as other configurable parameters selected for the task (e.g., the height of the robot's spine).
Choosing the robot's configuration can be challenging because of robot complexity, task complexity, geometric constraints, and kinematic constraints. Additionally, more than one configuration of the robot may be necessary to complete some tasks or more than one robot may be involved. For example, \citet{hawkins2014assistive} used two positions of a PR2 (a mobile manipulator made by Willow Garage) to reach and shave the entirety of a user's face (both sides) for a user in a wheelchair.

\begin{figure}[t]
\centering
\includegraphics[width=\columnwidth]{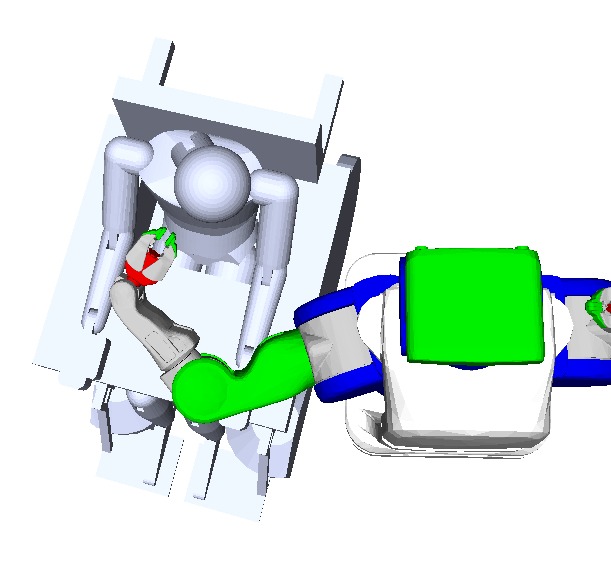}
\caption{TOC can select a single configuration for the PR2 to shave a person in a wheelchair, shown here. TOC takes advantage of the person's physical capabilities, such being able to rotate his or her neck.}
\label{fig:shaving_chair}
\end{figure}

Our approach to providing assistance to a user with a robot is to 
first find a good configuration for the robot from which it can perform the task.
In this work, we focus exclusively on the problem of selecting the robot configuration, leaving task performance out of scope. Performance of assistive tasks is addressed in many other works (see Section~\ref{ssec:rw_assistive_robots}).
The questions we ask are: how can the the robot select a good configuration, and what makes a configuration good? We address these questions in Section \ref{sec:method}.
With a good configuration, the robot is more likely to be able to
complete the task successfully.

We present a task-centric, optimization-based method to select one or more configurations for assistive robots, that we call task-centric optimization of robot configurations (TOC). TOC extends two previous works from \citet{kapusta2015task,kapusta2016optimization}, with changes to framework, formulation of terms used within the methods, and modeling of the user and environment. Additionally, we have improved the optimization search and performed additional evaluations of our method with thorough comparisons to other methods from the literature. 
TOC is suitable for quickly selecting one or more robot configurations for assistive tasks, including some activities of daily living (ADLs), that involve the robot moving a tool around a person's body. A task-centric approach is particularly applicable to assistive tasks where there may be 
important or common tasks that take place in environments that are known apriori. 
Key features of TOC are its selection of multiple configurations for a task, its task-centric approach, its explicit modeling of many task-specific parameters that are important to assistive tasks (possible because of the task-centric approach), its representations of robot dexterity, and its framework splitting offline and online computation.

TOC performs substantial offline computation to generate a function that 
can be applied rapidly online to select robot configurations based on current observations. These offline computations use
explicit models of the task, environment, and user.
Because offline modeling may be vulnerable to problems with error and mismatch between 
models and reality,
TOC uses two representations of robot dexterity that we developed, task-centric reachability (TC-reachability) and
task-centric manipulability (TC-manipulability), in its objective function to help it select robot
configurations that are implicitly robust to error. TOC searches the robot configuration space 
to maximize its objective function, a linear combination of TC-reachability and TC-manipulability, using a simulation-based, derivative-free optimization from literature, 
covariance matrix adaptation evolution strategy (CMA-ES).
Figures~\ref{fig:bathing_legs_autobed} and~\ref{fig:shaving_chair} show the configurations selected for the cleaning legs task for a user in bed and the shaving task for a user in a wheelchair, respectively.

\begin{figure*}[t]
\centering
\includegraphics[width=\textwidth]{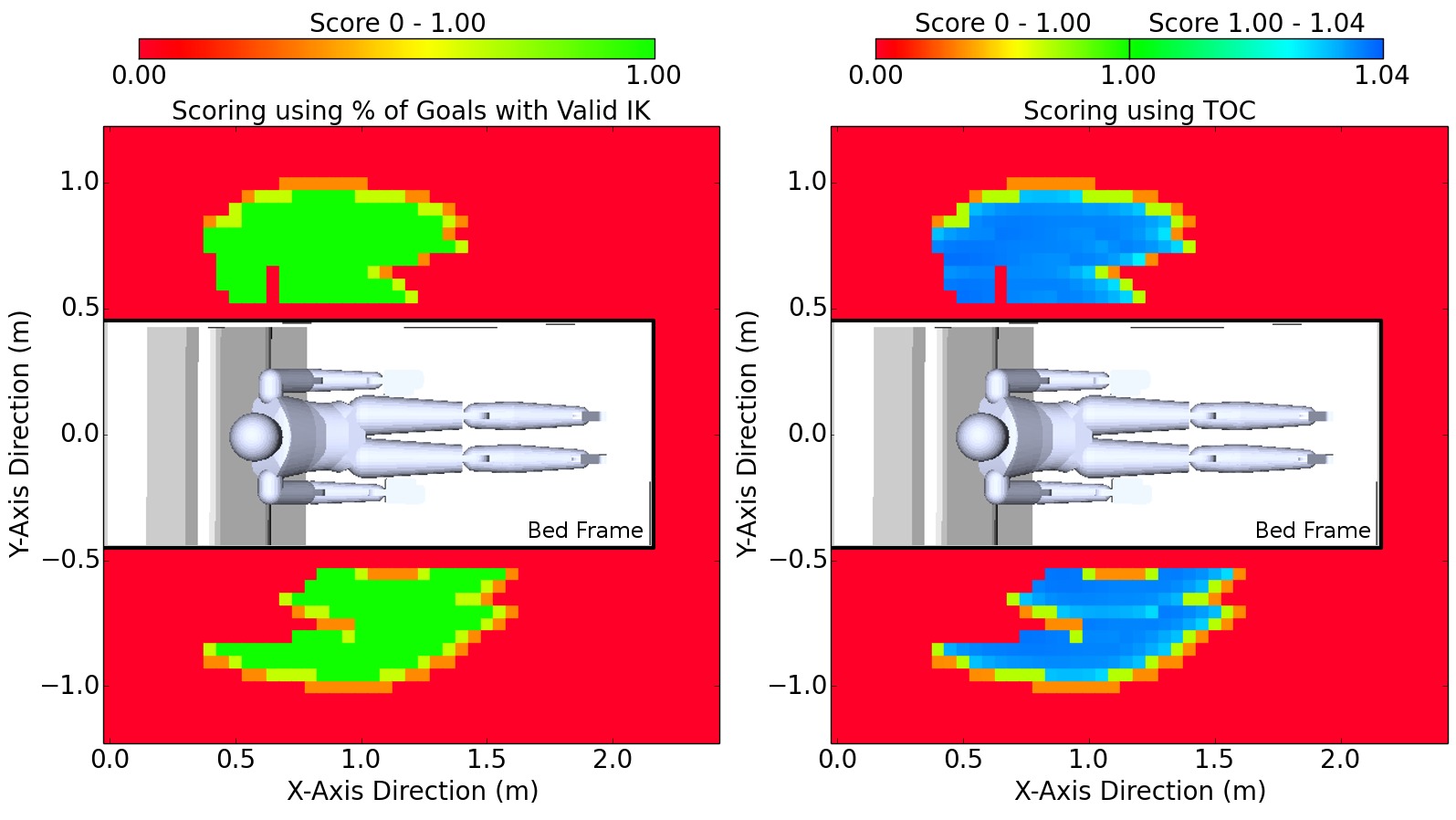}
\caption{TOC differentiates between robot configurations that have collision-free IK solutions to all goal poses. This differentiation allows TOC to better select good robot configurations. The figure visualizes two scoring methods, showing the score for discretized PR2 base poses. Z-rotation is sampled every \SI{45}{\degree} from \SI{0}{\degree} to \SI{360}{\degree}, and the best score is shown (left) using the percentage of goals with collision-free IK solutions and (right) TOC, for the mouth wiping task in the robotic bed environment, with the bed raised \SI{20}{\centi\metre} and at \SI{45}{\degree}.  The color represents the best score for that 2D position of the PR2. 0 means no goals and $\geq 1$ means all goals have collision-free IK solutions. }
\label{fig:toc_base_visualization}
\end{figure*}


We evaluated TOC in simulation with a PR2 assisting a user with 9 assistive tasks in both a wheelchair and a robotic bed. For our evaluations we implemented three baseline algorithms from literature. The first used an inverse-kinematics (IK) solver to select a configuration with a collision-free IK solution to all task-relevant goal poses. The second and third baselines used a capability map from \citet{zacharias2007capturing} to select a configuration with high capability score to the goal poses. The third baseline also checked that a collision-free IK solution existed for each goal pose. 
We ran Monte-Carlo simulations of pose estimation error and found that TOC's average success rate was higher than or comparable to baseline algorithms for each task.
TOC had an overall average success rate of 90.6\% compared to 50.4\% for the IK solver, 43.5\% for capability map, and 58.9\% for capability map with collision checking. Additionally, we provide evidence that TOC's objective function is positively correlate with robustness to error, and we demonstrate how TOC can be used to assist in designing environments to improve robotic assistance.




\section{Related Work}\label{sec:related_work}

\subsection{Representations of Robot Dexterity}
Many metrics have been developed to quantify the kinematic dexterity of robot manipulators. These metrics can be broadly divided into those that use the manipulator's Jacobian, $\boldsymbol{J}(\boldsymbol{q})$ \citep{spong2006robot} and those that do not. These metrics can also be divided into those that find global measures (a metric for the robot irrespective of joint configuration) and those that find local, configuration-dependent measures of dexterity. Global dexterity metrics are often used for robot design \citep{stocco1998matrix,hammond2011multi}. As we are focused on dexterity measures to assist in positioning existing robots, we will focus on discussing local metrics.
\citet{yoshikawa1984analysis} proposed the local Jacobian-based metric called measure of manipulability (or just, manipulability),  $w(\boldsymbol{q})$, shown in Equation~\ref{eq:manip}. 
\begin{equation}
 \begin{aligned}
 w(\boldsymbol{q}) = \sqrt{\text{det}(\boldsymbol{J}(\boldsymbol{q})\boldsymbol{J}(\boldsymbol{q})^T)}
 \end{aligned} \label{eq:manip}
\end{equation}
Geometrically, manipulability is proportional to the volume of
the manipulability ellipsoid of the manipulator, which is the volume of Cartesian space moved by the end effector for a unit ball of movement by the arm's joints. This metric can be useful
when assessing kinematic dexterity between different configurations of the same
robot. However, its scale and order dependencies make comparison between different robot morphologies challenging. 

Other dexterity measures were developed that address some of the issues of using manipulability \citep{klein1987dexterity}.
\citet{kim1991dexterity} proposed another local Jacobian-based metric that we refer to in this paper as kinematic isotropy, $\Delta(\boldsymbol{q})$, shown in Equation~\ref{eq:kinematic_isotropy}. 

\begin{equation}
 \begin{aligned}
 & \Delta(\boldsymbol{q})  = \frac{\sqrt[a]{\text{det}(\boldsymbol{J}(\boldsymbol{q})\boldsymbol{J}(\boldsymbol{q})^T)}}{(\frac{\text{trace}(\boldsymbol{J}(\boldsymbol{q})\boldsymbol{J}(\boldsymbol{q})^T)}{a})}
 \end{aligned} \label{eq:kinematic_isotropy}
\end{equation}

Kinematic isotropy uses the manipulability term (shown in Equation~\ref{eq:manip}) with an alteration to remove order dependency and divided by a term to remove scale dependency. 
Order is the size of the 
Cartesian space of interest. For a planar robot, the order would be three if translations are all in 2D in-plane and in-plane rotations are considered. For the case of tasks in 6D space (position and orientation), the order is six.  
Unlike manipulability, the values of kinematic isotropy always range from 0 to 1 they can be directly compared across robot platforms.
In our work we modified kinematic isotropy, adding a weighting term to create what we call joint-limit-weighted kinematic isotropy (JLWKI). We use JLWKI in our task-centric manipulability (TC-manipulability) defined in Section~\ref{ssec:configuration_scoring}. TC-manipulability represents the kinematic dexterity of the robot for a task (a set of goal poses) from a set of one or more positions of the robot.


An important limitation to some Jacobian-based measures of dexterity is their
ignorance of many relevant features of the workspace, such as joint limits and collisions. The manipulability ellipsoid calculated from the Jacobian suggests that the end effector can move in ways
that may be constrained by joint limits.
Many researchers have proposed various ways to include these features (joint limits: \citet{tsai1986workspace, chan1995weighted}; velocity limits: \citet{lee1997study}; torque limits: \citet{hammond2009improvement}) into weighting terms. 
\citet{vahrenkamp2012manipulability} created what they called an extended manipulability 
measure by modifying the Jacobian to include weights on joint limits, on proximity to self-collision, and on proximity to collision
with the environment. 
Many of these methods apply their weighting terms directly to manipulability or indirectly by modifying the Jacobian, which is used in manipulability. JLWKI differs from other measures of dexterity, using a distinct weighting function on joint limits and applying it to kinematic isotropy. 
JLWKI, does not include costs on proximity to collision because we found that calculating these costs increases the computation time of TOC excessively.

\citet{zacharias2007capturing} introduced a method for representing manipulator
dexterity without using Jacobians, which they use to 
score the workspace of a robot, creating what they call a capability map. To create the capability map they discretize space around the robot into 3D points and discretize the range of possible orientations around each 3D point. The capability score (also known as reachability score) is the number of orientations for which the robot has a valid
IK solution. A way to interpret the meaning of this score is, if a goal pose is located at the 3D location, the capability score is similar to the probability that the manipulator can achieve the pose.

\subsection{Selecting Robot Configurations for Mobile Manipulation}\label{ssec:rw_robot_configurations}
Prior research has investigated how to select configurations for a mobile robot. 
A common method is to address the problem using IK solvers \citep{diankov2010automated, beeson2015trac, kumar2010optimization, kdl-url}. The entire kinematic chain from end effector to the robot's base location may be solved using IK \citep{gienger2005task, grey2016humanoid}. Alternatively, sampling-based methods  may be used to find robot base poses that have valid IK solutions, often as part of motion planning \citep{elbanhawi2014sampling, stilman2005navigation, lindemann2005current, garrett2015backward, diankov2008bispace}. 

By relying solely on IK to ensure that the robot can reach the goals, these methods are dependent on accurate models. Many of these methods are fast, but may fail if there is modeling or state estimation error. Like these methods, TOC uses a sampling-based search to find a robot configuration with valid IK solutions. However, there are many such robot configurations, and they cannot be distinguished using only IK, as shown in Figure~\ref{fig:toc_base_visualization}(left). All locations in green have collision-free IK solutions to all goals, but some may result in higher success rates than others. TOC uses task-centric manipulability to differentiate those configurations, as shown in Figure~\ref{fig:toc_base_visualization}(right). We show in Section~\ref{ssec:evaluation_scoring} that higher TOC score is correlated with improved performance for configurations that have collision-free IK solutions to all goals. Additionally, TOC can find more than one robot configuration for a task. We implemented a standard IK sampling-based method as a baseline for comparison, as described in Section~\ref{ssec:evaluation_vs_baseline}. 

A body of work is based on the capability map 
from \citet{zacharias2007capturing}.
Capability-map-based methods are robot-centric and task-agnostic; they are generated offline for the robot's manipulator and applied to tasks online. They typically select the robot base position by overlapping the capability map with end effector goal poses and maximizing the 
average capability score \citep{zacharias2009using, porges2014reachability, leidner2014object}. 

\citet{dong2015orientation} altered the capability map 
by creating an orientation-based capability map and extending the map
for tools on the robot's end effector. 
In contrast with our method, existing capability-map-based methods do not consider collisions with the environment in their offline computations because they do not model the environment. Collisions are only considered at runtime to eliminate robot base locations in collision or that lack collision-free IK solutions. Simply selecting the robot base location with highest capability map score is fast, but searching for a collision-free location can take more time. Capability-map-based methods also only find a single location for a robot for a task, but our method can find multiple robot configurations. We implemented two capability-map-based methods as baselines for comparison, based on \citet{zacharias2009using}, as described in Section \ref{ssec:evaluation_vs_baseline}. 

Another body of work extends the capability map by inverting it, creating an inverse-reachability map. While a capability map scores end effector poses with respect to a robot base pose, the inverse-reachability map scores robot base poses with respect to an end effector pose. As with the capability map, the inverse-reachability map is generated offline for the robot's manipulator and can be used quickly online. For an end effector 3D position, discretized robot base poses are scored based on the capability map score to that 3D position. The inverse-reachability map is used to rapidly sample a robot base position that can reach a set of goal end effector poses \citep{diankov2008openrave, burget2015stance}. 
\citet{vahrenkamp2013robot} used an alternative representation of the robot's dexterity from their previous work \citep{vahrenkamp2012manipulability} that uses 6D poses in the workspace. They invert that workspace representation to create what they call an Oriented Reachability Map (ORM). ORM, like the inverse-reachability map,
scores robot base poses based on the extended manipulability measure of each 6D pose in the manipulator's discretized workspace. When searching for a robot base pose for a task, they sample in series from the ORM using the map's score as a sampling weight. If the sampled robot base pose has collision-free IK solutions to the goal end effector poses, they use that base pose; if not they re-sample.
They propose methods to incorporate task-specific information in their extended manipulability measure (and thus into the ORM), and to calculate the ORM map online through what they call lazy-ORM. 
Inverse-reachability-based methods and ORM differ from our work in a few ways. These methods are typically used in task-agnostic and robot-specific ways to facilitate applications of the robot to new tasks. Notably, these methods only find a single location for a robot for a given task, rather than multiple robot configurations. TOC also explicitly models 
features and parameters of the environment and user that may be important to assistive tasks. Details
on this modeling is found in Sections~\ref{ssec:environment_modeling} and~\ref{ssec:user_modeling}.

Most previous task-centric methods use simulation of the task, with explicit error modeling, to evaluate robot base poses. 
\citet{hsu1999placing} presented a task-specific method for selecting a place for 
an industrial robot
manipulator to perform a series of tasks amidst clutter. They used 
randomized path planners to generate collision-free paths for the arm and they 
randomly perturb the robot position to find positions from which the tasks can be
performed quickly.  
\citet{stulp2009learning} present a task-centric method for 
finding areas in which to 
place a mobile manipulator where it can successfully perform 
a grasping task. They use Monte-Carlo simulation of 
error in the location of the object to be grasped to find base positions 
with high success rates. They simulate performance of the entire task
including, navigation, motion planning and motion execution. For real-time 
base position selection, they convolve uncertainty in robot location with base
position scores to provide an area of high-success probability. They used their method to select a 2D position
of the robot base for a grasping task. These task-centric methods
that explicitly model error and fully simulate task performance
have only been used to select a few degrees of freedom in static environments, and can only select a single robot configuration for a task. In contrast, TOC uses faster, simpler simulation and implicitly handles error. TOC selects more degrees of freedom in configurable environments, and, again, can select multiple robot configurations for a task.

\subsection{Human-robot Proxemics}\label{ssec:rw_proxemics}
Several bodies of work have examined the proxemics of human-robot interactions \citep{walters2011long, walters2009empirical, mumm2011human, takayama2009influences, walters2005influence}. 

Proxemics is the study of the spatial requirements of humans (e.g., the amount of space that people feel it necessary to set between themselves and others). These works look at acceptable interpersonal distances
between humans and robots in social settings. 
Various works have used the concepts of human-robot proxemics to inform a robot
when performing tasks. These works couple task performance concepts with scoring methods
based on proxemics to select base positions and paths for the robot and item handover locations \citep{sisbot2010synthesizing, mainprice2011planning, sisbot2006mobile, sisbot2007human}. Proxemics might suggest that placing the robot in front of the person at some minimum distance is preferred over other locations.

\citet{kruse2013human} present a thorough survey of human-aware robot navigation.
In contrast, TOC does not consider proxemics or social
factors; it instead focuses on kinematic aspects of the task. 
However, inclusion of additional terms in TOC's objective function to include user comfort
and proxemics is possible. 
While proxemics is often used to consider navigation problems, TOC focuses exclusively on selecting the configuration for the task, which may be the goal pose for navigation.
 
\subsection{Assistive Robots}\label{ssec:rw_assistive_robots}
Researchers have investigated the use of mobile manipulators as assistive devices \citep{dario1999movaid, schaeffer1999care, graf2009robotic, bien2004integration, jain2010assistive, hawkins2014assistive}.  

We seek to further empower assistive mobile manipulators by autonomously selecting configurations from which they can better provide assistance. This autonomy can improve task performance and decrease cognitive workload for teleoperated assistive systems, as from \citet{grice2016assistive}. 
In this work, we have used a model of a robotic bed that matches Autobed, a robotic modified hospital bed from \citet{grice2016resna}. We have shown how TOC can optimize the configuration of the bed, allowing improved assistance from a mobile manipulator as part of a collaborative assistive system, as from \citet{kapusta2016collaboration}. This capability is demonstrated in our evaluations in the robotic bed environment in Section~\ref{ssec:evaluation_vs_baseline}.

\begin{figure*}[t]
\centering
\includegraphics[width=12cm]{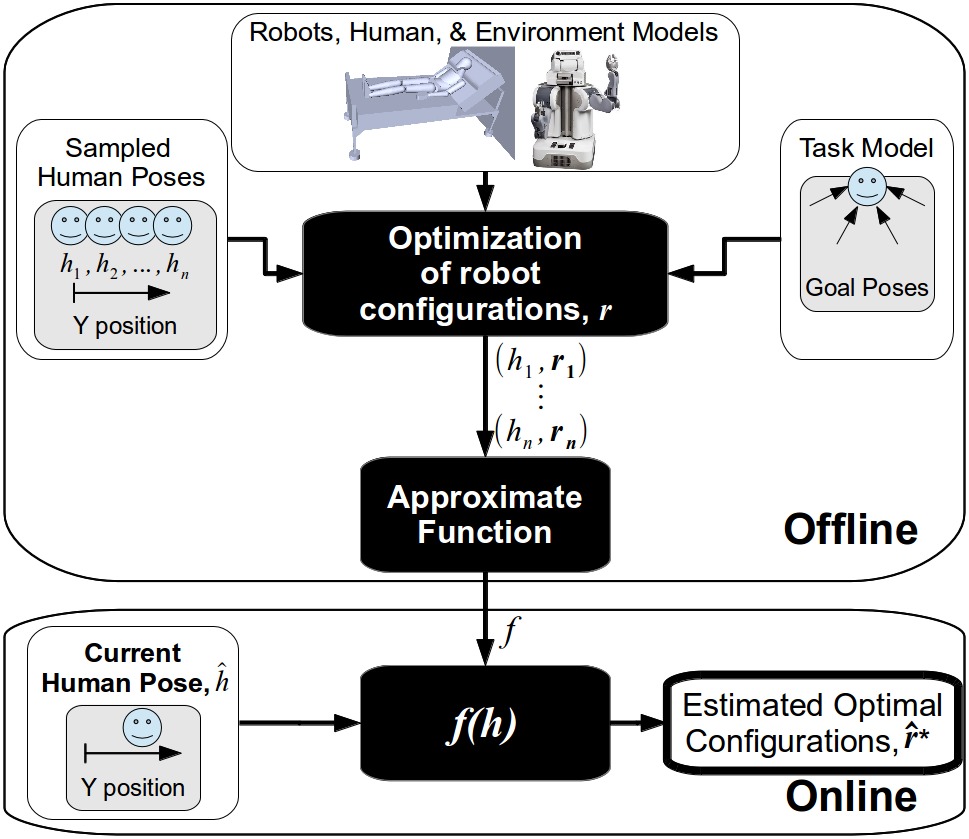}
\caption{The Framework used in TOC. The offline portion of TOC takes as input task-relevant models
and samples of the uncontrollable parameters and outputs optimized robot configurations. It then approximates a function that is used online to estimate the optimal robot configurations given the current, observed uncontrollable parameters.}
\label{fig:toc_framework}
\end{figure*}



\section{Task-centric Optimization of Robot Configurations (TOC)}\label{sec:method}
As mentioned in Section~\ref{sec:introduction}, key features of task-centric optimization of robot configurations (TOC) are its task-centric approach, representations of robot dexterity, selection of multiple configurations for a task, and framework that splits offline and online computation. 
TOC is suitable for situations when tasks and environment layouts are 
known beforehand and we would like 
to configure the robot for these tasks such that the robot is successful despite 
variations between models and reality. By taking a task-centric approach, TOC is 
able to use task-specific knowledge, such as explicit modeling of task-relevant parameters, to better select configurations.
We will first explain the goal of TOC. Afterwards we describe the nomenclature used in the remainder the paper.
We then explain the framework of TOC, details of its features, and specifics of our implementation.

\subsection{TOC Goal: Selecting Good Configurations}\label{ssec:good_config}
The goal of TOC is  
to select a good set of one or more configurations for a robot to perform a
task without additional adjustments.  
But what constitutes a \emph{good} robot configuration?
In this paper we use the term \emph{robot configuration} as a more
general term for the pose of the robot's base, so it can include additional relevant parameters.
For example, for a PR2, the robot configuration might be the position and orientation of the 
robot's mobile base as well as the z-axis spine height of the robot. If the PR2 were operating in a 
room with a robotic bed, the degrees of freedom (e.g., the height of the bed) of the bed could be included in the robot configuration. 
We consider robot configurations in sets that can be of cardinality 1 or greater; the robot can complete the entire task by adopting all configurations in the set in any order.
With a \emph{good} set of configurations, the robot is more likely to be able to
complete the task successfully. 
We judge the robot's ability to perform the task from a set of 
robot configurations with one measure: if it can reach all goal poses collision-free with its end effector. 
Various forms of error, such as modeling error or state estimation error, may cause the robot to be unable to perform the task.
Because the robot does not know how the modeling and state estimation error will manifest apriori, from a good set of robot configurations, the robot should be able to
perform the task despite such error. 

\subsection{Nomenclature}\label{ssssec:nomenclature}\hspace*{\fill} \\
\begin{tabularx}{\columnwidth}{p{1.0cm} X}
$c$: & A task identifier 
\end{tabularx}
\begin{tabularx}{\columnwidth}{p{1.0cm} X}
$N_c$: & The number of goal poses for task $c$
\end{tabularx}
\begin{tabularx}{\columnwidth}{p{1.0cm} X}
$\boldsymbol{q}$: & A joint configuration of the robot arm. $\boldsymbol{q} \in \mathbb{R}^n$, where $n$ is the number of DoF of the arm
\end{tabularx}
\begin{tabularx}{\columnwidth}{p{1.0cm} X}
$q_{i}$: & The value for joint $i$ in joint configuration $\boldsymbol{q}$, $q \in \boldsymbol{q}$
\end{tabularx}
\begin{tabularx}{\columnwidth}{p{1.0cm} X}
$\boldsymbol{q}^{-}, \boldsymbol{q}^{+}$: & A list of the minimum and maximum values, respectively, for the joints of a robot's arm.
\end{tabularx}
\begin{tabularx}{\columnwidth}{p{1.0cm} X}
$q^{-}_{i},q^{+}_{i}$: & The minimum and maximum values, respectively, for joint $i$ of a robot's arm.
\end{tabularx}
\begin{tabularx}{\columnwidth}{p{1.0cm} X}
$\boldsymbol{r}$: & A set of robot configurations of cardinality $\geq 1$,  $\boldsymbol{r}=\{r_1, r_2, ..., r_n\}$, where
$n$ is the number of robot configurations in set $\boldsymbol{r}$ 
\end{tabularx}
\begin{tabularx}{\columnwidth}{p{1.0cm} X}
$\boldsymbol{\hat{r}}^{*}$: & The estimated optimal robot configurations given current observations, the output of the online portion of TOC.
\end{tabularx}
\begin{tabularx}{\columnwidth}{p{1.0cm} X}
$\boldsymbol{h}$: & The set of uncontrollable parameters, discretized into $\{h_1, h_2, h_3, ... \}$ 
\end{tabularx}
\begin{tabularx}{\columnwidth}{p{1.0cm} X}
$\hat{h}$: & The uncontrollable parameters observed and estimated at run-time, the input to the online portion of TOC.
\end{tabularx}
\begin{tabularx}{\columnwidth}{p{1.0cm} X}
$\boldsymbol{b}$: & The set of free parameters, discretized into $\{b_1, b_2, b_3, ... \}$ 
\end{tabularx}
\begin{tabularx}{\columnwidth}{p{1.0cm} X}
$\boldsymbol{x}$: & Set of position and orientation end effector goal poses $x \in \mathbb{R}^6$. $\boldsymbol{x}$ depends on $c$, $h$, and $b$, but we omit those for simplicity in writing. $\boldsymbol{x}=\{x_{1}, x_{2}, ..., x_{N_c}\}$.  
\end{tabularx}
\begin{tabularx}{\columnwidth}{p{1.0cm} X}
$\boldsymbol{s}_{r, x}$: & Set of IK joint configuration solutions to goal $x$ from \emph{robot configuration} $r$, $\boldsymbol{s}_{r, x}=\{\boldsymbol{q}_1, \boldsymbol{q}_2, ..., \boldsymbol{q}_n\}$, where $n$ is the number of IK solutions
\end{tabularx}
\begin{tabularx}{\columnwidth}{p{1.0cm} X}
$a$: & The order of the robot arm. In our case, 6. 
\end{tabularx}
\begin{tabularx}{\columnwidth}{p{1.0cm} X}
$\boldsymbol{J}(\boldsymbol{q})$ & The Jacobian of the arm in joint configuration $\boldsymbol{q}$ 
\end{tabularx}
\begin{tabularx}{\columnwidth}{p{1.0cm} X}
$\Delta(\boldsymbol{q})$: & The kinematic isotropy for the arm in joint configuration $\boldsymbol{q}$ 
\end{tabularx}
\begin{tabularx}{\columnwidth}{p{1.0cm} X}
$f$: & A function that takes $\hat{h}$ as input and outputs $\boldsymbol{\hat{r}}^{*}$. TOC generates $f$ offline and applies it online.
\end{tabularx}

\subsection{Framework}\label{ssec:toc_framework}
Figure~\ref{fig:toc_framework} shows the framework of TOC for our implementation
described in Section~\ref{sec:evaluation} for a person on a robotic bed.
TOC performs most of its computation offline to approximate a function
that can be used online to select robot configurations for a task. 
The optimization takes as input task-relevant models (e.g., task, robot, user, and environment models) and a sample of the uncontrollable parameters (e.g., the position of the person on the bed), $h$. It outputs an optimized robot configuration, $\boldsymbol{r}$, for that $h$.
The inputs and outputs of the optimization are used
to approximate the function, $f$ that takes as input at run-time the observed estimated uncontrollable parameters, $\hat{h}$ and outputs the estimated optimal configurations, $\boldsymbol{\hat{r}}^{*}$.

\subsection{Task Modeling}\label{ssec:task_modeling} 
\begin{figure}[t]
\centering
\includegraphics[height=4cm]{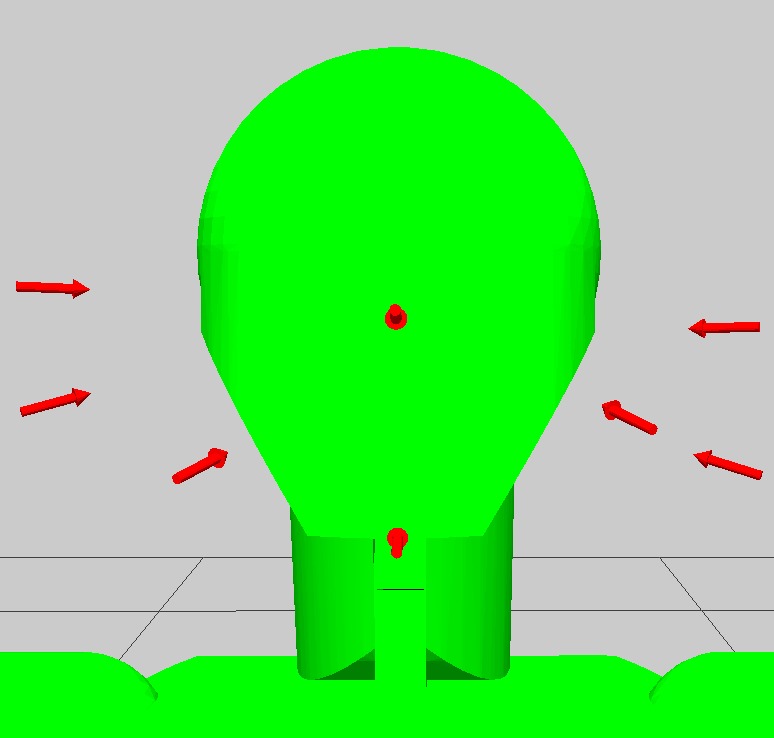}
\hspace{1mm}
\includegraphics[height=4cm]{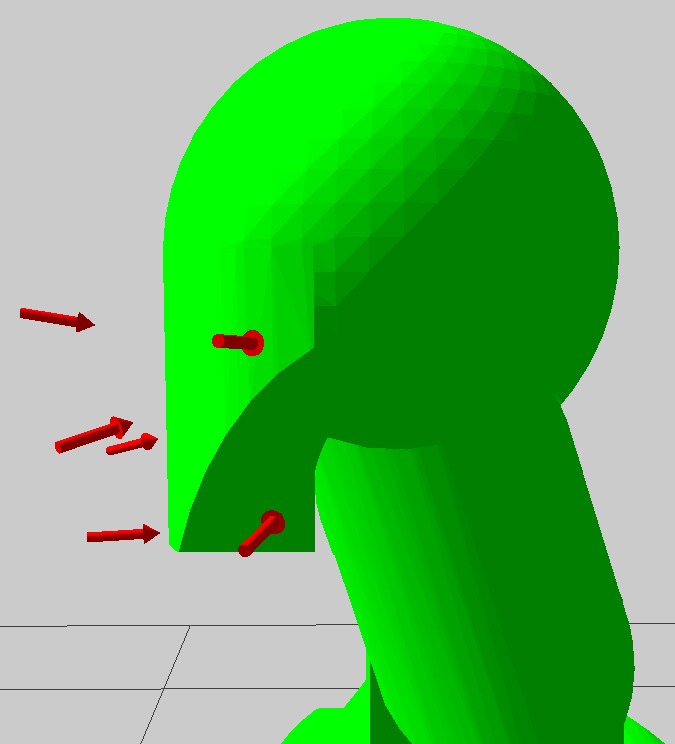}
\caption{The manually selected goal poses for the shaving task. Each arrow represents a position and orientation, 6-DoF end effector goal pose, with respect to the head. This shows views from the front and side.
}
\label{fig:shaving_task_goals}
\end{figure}

Our aim with task modeling is to create a representation that allows TOC
to efficiently evaluate a robot's ability to perform a task.
There are many tasks that consist of manipulation of small objects or tools around a person's body, that we expect can be well modeled by a set of goal poses (Cartesian positions and quaternion orientations) with respect to relevant reference frames.
We manually model each task as a sparse set of poses for the robot's
end effector. For example, 
Figure~\ref{fig:shaving_task_goals} shows the eight goal poses with respect to the person's head that make up our model of a shaving task. 
We assume that if the robot can reach all goal poses, it is likely to be able 
to perform the task. However, we expect
there to be differences between the models and the real tasks. We consider these discrepancies
to be a form of modeling error that TOC accounts for when selecting robot configurations.



\subsection{Environment Modeling}\label{ssec:environment_modeling}
Using its environment model, TOC finds robot configurations that avoid collision and unwanted interaction with obstacles, such as a bedside table or walls. TOC can use different resolutions for its environment model depending on the needs of the task. 
A room could be be represented simply as a wall behind the bed, as shown in Figure~\ref{fig:bathing_legs_autobed}, or it could contain models of furniture and other potential obstacles. The resolution of each object model could range from a block to a detailed mesh. 
TOC uses three types of objects to model the environment:
\begin{itemize}
\item Fixed objects
\item Controllable movable/configurable objects
\item Uncontrollable movable/configurable objects
\end{itemize}
TOC treats fixed objects as static obstacles in the world, to be avoided by the robot. We add the configuration of the controllable objects
to the robot configuration space that TOC optimizes. An example of a controllable object is a adjustable bed. For uncontrollable objects, TOC selects a robot configuration for a sample of the possible configurations of the objects. For each sample of its configuration, the object is considered static. An example of an uncontrollable object is a nightstand that is not movable by the robot, but could be moved somewhere by a person prior to the robot starting the task.
Using controllable and uncontrollable movable objects, TOC can suggest alterations to the environment 
that may improve task performance. Uncontrollable movable objects can also be used to generate robot configurations for possible states of the environment. This may be beneficial for environments, such as hospitals, where there are a few possible room layouts, but the robot may not know which layout will be relevant until it reaches the room.

Because TOC does not include a cost in proximity to collision between the robot and the environment in its objective function, we include a margin of safety in the environment model by expanding the environment model (we used $\sim$\SI{3}{\centi\metre} in our evaluations). This safety margin reduces the risk of collision in the case of model or state estimation error, without having to explicitly include closeness to collision in the objective function.

\subsection{User Modeling}\label{ssec:user_modeling}
TOC's user model can be customized for a user
to better locate relevant parts of the body, and to allow more accurate collision-checking. 
In our evaluation of TOC we used a mesh model of a human designed around a 50~percentile male from \citet{human_measurement}, shown in Figure~\ref{fig:bathing_legs_autobed}.
TOC uses three types of parameters for the person's configuration:
\begin{itemize}
\item Environment-driven parameters
\item Uncontrollable parameters
\item Free parameters
\end{itemize}
Environment-driven parameters are set according to the state of the
environment. For a chair, the user's body would be in a seated configuration. For a flat bed,
the user's body would be in a supine configuration. For a bed with an adjustable back rest, the user's
configuration would depend on the angle of the back rest. Uncontrollable parameters are treated similar 
to uncontrollable movable objects. TOC selects a robot configuration for a sample of the 
uncontrollable parameter. An example of an uncontrollable parameter is the position of the user on the 
bed, if the robot is unable to shift the body on the bed. Free parameters are used by TOC freely
without including it in the robot configuration. An example of a free parameter used is the user's neck 
rotation. 
Figure~\ref{fig:shaving_chair} shows a configuration of the PR2 that takes advantage of the user's neck rotation to reach all goals for the shaving task in a wheelchair.

Just as with the environment model, we include a margin of safety in the human model by expanding the model.

\subsubsection{Additional User Customization}
TOC can consider additional customizations for the user's needs or preferences.
For example, a user may prefer certain angles of
the bed's head rest for feeding tasks. This preference can be represented as limitations or costs 
on the robot's configuration space.

\subsection{Configuration Scoring}\label{ssec:configuration_scoring}
Implicitly handling variation and error is a key aspect of TOC, because its heavy computation is 
performed offline for models that may differ from reality. 
TOC uses two metrics that we have developed to estimate how well the robot will be able to perform the task from a set of configurations: task-centric reachability (TC-reachability) and task-centric manipulability (TC-manipulability). 


\subsubsection{Task-centric Reachability}

Task-centric Reachability (TC-reachability), $P_R$, is the percentage of goal poses to which the robot can find a collision-free IK solution from robot configurations, $\boldsymbol{r}$, for a task $c$ and uncontrollable parameters $h$, as shown in Equation~(\ref{eq:reachability}). 
\begin{equation}
 \begin{aligned}
  && P_{R}(\boldsymbol{r},c,h)  = (\frac{1}{N_c}) \sum\limits_{k=1}^{N_c} \operatorname*{max}_{r \in \boldsymbol{r}, b \in \boldsymbol{b}} W(r, x_{k}) ,
 \end{aligned} \label{eq:reachability}
\end{equation}
where
\begin{equation}
 \begin{aligned}
 & W(r, x_{k}) = 1 &\forall \boldsymbol{s}_{r, x_{k}} \neq \emptyset,\\
 \text{and}\\
 & W(r, x_{k}) = \text{ }0 & \forall \boldsymbol{s}_{r,x_{k}} = \emptyset.
 \end{aligned} \label{eq:toc_f_function}
\end{equation}
\noindent
Recall that $\boldsymbol{x}$ depends on $c$, $h$, and $b$, but we omit those for simplicity in writing and $N_c$ is the number of goal poses for task $c$. Note that $\boldsymbol{s}_{r, x_{k}} \neq \emptyset$ means that the IK solver can find a collision-free solution to the goal pose $x_{k}$ from robot
configuration, $r$.

TC-Reachability is related to using an IK solver with collision checking, but with the additional functionality of evaluating sets of robot configurations.

\subsubsection{Task-centric Manipulability}

Task-centric Manipulability (TC-manipulability), $P_M$, is related to the average kinematic dexterity of the arm when reaching the goal poses. It is defined here differently from our previous works, such as from \citet{kapusta2015task}. 

\begin{figure}[t]
\centering
\includegraphics[width=7.5cm]{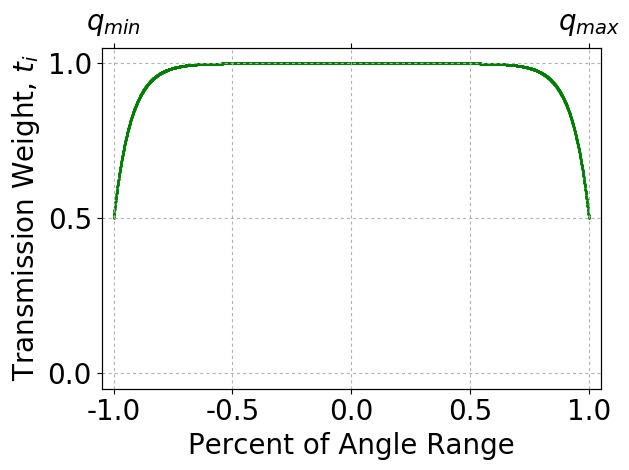}
\caption{A plot of the joint-limit weighting function ranging from maximum joint value to the minimum joint value.}
\label{fig:joint_limit_weighting}
\end{figure}

TC-manipulability score is based on kinematic isotropy \citep{kim1991dexterity}, shown in Equation~(\ref{eq:kinematic_isotropy}). 
Kinematic isotropy only considers the Jacobian of the arm in a configuration, ignoring potentially relevant properties of the robot arm, such as joint limits. When at a joint limit, the arm cannot move in one direction, effectively halving the movement of that joint.
\citet{hammond2009improvement} used torque-weighted global isotropy index and torque-weighted kinematic isotropy to estimate the dexterity of a robotic arm given joint torques and torque limits. \citet{vahrenkamp2012manipulability} investigated configuration-based weighting functions to create what they call an augmented Jacobian that they use in manipulability.
We have similarly modified kinematic isotropy to consider joint limits by scaling the manipulator's Jacobian by an $n \textnormal{x} n$ diagonal joint-limit-weighting matrix $\boldsymbol{T}$, defined in Equation~\ref{eq:toc_weighting_matrix}, where $n$ is the number of joints of the manipulator.
\begin{equation}
 \begin{aligned}
 & \boldsymbol{T}(\boldsymbol{q}, \boldsymbol{q}^{-}, \boldsymbol{q}^{+}) & = && 
 \begin{bmatrix} 
 t_1 & 0 & 0 \\[0.0em]
 0 & \ddots & 0 \\[0.0em]
 0 & 0 & t_n \\[0.3em]
 \end{bmatrix}
 \end{aligned}\label{eq:toc_weighting_matrix} 
\end{equation}
\noindent
$t_i$ in $\boldsymbol{T}$ is defined as
\begin{equation}
 \begin{aligned}
& t_i  =  1-\eta^{\kappa} \label{eq:toc_updated_t}\\
 \text{where}\\
  &\kappa = \frac{q^{r}_{i} -\lvert q^{r}_{i} - q_{i} + q^{-}_{i}\rvert}{\zeta q^{r}_{i}}+1 \\
  \text{and}\\
 &q^{r}_{i} = \frac{1}{2}(q^{+}_{i} - q^{-}_{i}).
 \end{aligned} 
\end{equation}
We set $t_i=1$ for infinite roll joints. 
The variable 
$\eta$ is a scalar value that determines the maximum penalty incurred when joint $q_i$ approaches $q^{+}_{i}$ or $q^{-}_{i}$ and $\zeta$ determines the shape of the penalty function. We used a value of $0.5$ for $\eta$ and $\frac{1}{20}$ for $\zeta$.  
This weighting function and the values for $\eta$ and $\zeta$ were selected to halve the value
of the kinematic isotropy at joint limits, have little effect in the center of the joint
range, to begin exponentially penalizing joint values beyond 75\% of the range, and to 
operate as a 
function of the percentage of the joint range.
Fig. \ref{fig:joint_limit_weighting} shows the value of $t_i$ as a function of the joint value as a percentage of its joint range. 

We then define joint-limited-weighted kinematic isotropy (JLWKI) as
\begin{equation}
 \begin{aligned}
 & \text{JLWKI}(\boldsymbol{q}) =  \frac{\sqrt[a]{\text{det}(\boldsymbol{J}(\boldsymbol{q})\boldsymbol{T}(\boldsymbol{q}, \boldsymbol{q}^{-}, \boldsymbol{q}^{+}) \boldsymbol{J}(\boldsymbol{q})^T)}}{(\frac{1}{a})\text{trace}(\boldsymbol{J}(\boldsymbol{q})\boldsymbol{T}(\boldsymbol{q}, \boldsymbol{q}^{-}, \boldsymbol{q}^{+})\boldsymbol{J}(\boldsymbol{q})^T)}.
 \end{aligned} \label{eq:jlwki}
\end{equation}

We use a function, $F$, to find the maximum value of $\text{JLWKI}(\boldsymbol{q})$ for robot configuration $r$ and goal pose $x_k$, where
\begin{equation}
 \begin{aligned}
 & F(r, x_{k}) = \operatorname*{max}_{\boldsymbol{q} \in \boldsymbol{s}_{r,x_{k}}}\text{JLWKI}(\boldsymbol{q}) &\forall \boldsymbol{s}_{r, x_{k}} \neq \emptyset,\\
 \text{and}\\
 & F(r, x_{k}) = \text{ }0 & \forall \boldsymbol{s}_{r,x_{k}} = \emptyset.
 \end{aligned} \label{eq:toc_f_function}
\end{equation}

We finally define TC-manipulability, $P_{M}$, as
\begin{equation}
 \begin{aligned}
 & P_{M}(\boldsymbol{r},c,h)  = (\frac{1}{N_c}) \sum\limits_{k=1}^{N_c} \operatorname*{max}_{r \in \boldsymbol{r},b \in \boldsymbol{b}} F(r, x_{k}). 
 \end{aligned} \label{eq:tcs_manipulability}
\end{equation}

\subsection{Optimization}\label{ssssec:optimization}
TOC's optimization takes as input task-relevant models for task, $c$, and a sample of the uncontrollable parameters, $h$. It outputs an optimized set of robot configurations, $\boldsymbol{r}$. TOC runs this optimization for samples of the uncontrollable parameters for each task. 
TOC searches the robot configuration space to maximize its objective function, a linear combination of 
TC-Reachability and TC-Manipulability, shown in Equation~\ref{eq:objective_function}. 

\begin{equation}
 \begin{aligned}
 & \operatorname*{arg\,max}_{\boldsymbol{r}_i} &  \alpha P_{R}(\boldsymbol{r}_i, h_i, c) + \beta P_{M}(\boldsymbol{r}_i, h_i, c)
 \end{aligned} \label{eq:objective_function}
\end{equation}

Both TC-reachability and TC-manipulability range from 0 to 1, allowing them to be directly compared in the optimization. We selected a value of 1 for $\alpha$. We chose to define $\beta$ as
\begin{equation}
 \begin{aligned}
 & \beta(\boldsymbol{r})  = (0.1)(0.95)^{n-1}
 \end{aligned} \label{eq:tcs_manipulability}
\end{equation}
where $n$ is the cardinality of $\boldsymbol{r}$. In our implementation of TOC the cardinality of $\boldsymbol{r}$ was 1 or 2. These definitions of $\alpha$ and $\beta$ emphasize the importance of reaching goals over being able to reach around goals, and includes a small penalty in the objective function's value for using more configurations. 
There are often many configurations that can reach all goals. TC-manipulability is used to differentiate between these configurations. 
As an example, we compare TOC to a standard method from literature: using an  IK solver with a collision checker to find a robot configuration that can reach all goals. Figure~\ref{fig:toc_base_visualization} shows the difference in scoring between using the existence of IK solutions for scoring and using TOC for scoring for a task for a user in bed. Figure~\ref{fig:toc_base_visualization} (left) 
shows that many poses of the robot's base have the same score, each having collision-free IK solutions to all goals. Figure~\ref{fig:toc_base_visualization} (right)
shows scoring using TOC, where TC-manipulability allows additional differentiation between robot base poses that can reach all goals. Higher TC-manipulability is correlated with mean accuracy (percentage of goals that are reachable), as we show in Section~\ref{ssec:evaluation_scoring}.

\subsubsection{Search Method}
The space of the objective function can be highly nonlinear and challenging to search. There are several derivative-free, simulation-based optimization methods that could be applied to this problem. A simple method would be to uniformly sample the space and select the configuration with highest objective function value. However, we found that uniform sampling had difficulty finding good configurations for tasks where the solution space was small. 
\Urlmuskip=0mu plus 2mu
\citet{tan2011articulated} used Covariance Matrix Adaptation Evolution Strategy~(CMA-ES) to design a controller for 
articulated bodies moving in a hydrodynamic environment,
which inspired our use of CMA-ES (from \url{https://pypi.python.org/pypi/cma}) for our optimization. 
\Urlmuskip=0mu plus 0mu
We used a heuristic when both $P_{R}$ and $P_{M}$ are zero that pushes the search toward configurations that may have non-zero $P_{R}$ and $P_{M}$. 
All values of the heuristic are less than 0. 
 

\subsection{Approximate Function}\label{ssssec:function_approximation}
Offline, TOC approximates a function that takes as input an estimation of the uncontrollable parameters, $\hat{h}$, such as the location of the person on the bed, and outputs the estimated optimal robot configurations, $\boldsymbol{\hat{r}}^{*}$. At run-time, TOC applies this function to the observed, estimated uncontrollable parameters. For this paper we used K-nearest neighbor (K-NN) with $K=1$ (hence, 1-NN) as the function, $f$. We trained the 1-NN algorithm with a set of ($h$, $r$) pairs and it  returns as $\boldsymbol{\hat{r}}^{*}$ the $r$ for the $h$ that is closest to $\hat{h}$. In our implementations of TOC we trained the 1-NN on fewer than 20 ($h$, $r$) pairs for each task and found that the 1-NN would return $\boldsymbol{\hat{r}}^{*}$ in less than $1$~second.

\section{Evaluation}\label{sec:evaluation}
\subsection{Implementation}\label{ssec:implementation}
We manually created models for 
9 assistive tasks: shaving, feeding, wiping the mouth, cleaning both arms, cleaning both legs, and scratching 
the left/right upper arm, and left/right knee (each 
scratching task was considered separately). 
Previous work has noted that these types of tasks may be useful for those with severe motor impairments \citep{chen2013robots}.
As described in Section \ref{ssec:task_modeling}, task models consisted of a set of
goal poses, each of which was a position and orientation goal for the robot's end effector. We  defined each goal pose with respect to 
a relevant reference frame (e.g., the head for shaving, or the shoulder for scratching the upper arm), so they
move appropriately as the model parameters change (e.g., the height of the bed).
We chose these tasks as representative of various activities for which a 
robot like the PR2 may be able to provide assistance to a user with motor impairments. 

For example, 
Figure~\ref{fig:shaving_task_goals} shows the eight goal poses with respect to the person's head that we selected to model the shaving task. 
For simplicity, we limited tasks to one-handed tasks and 
used only the robot's left arm in our evaluations. In our implementation we allowed TOC to search for sets of robot configurations of cardinality 1 or 2. When exploring multiple robot
configurations for a task, we assume the robot can move from one configuration to another.

\begin{figure*}[!t]
\centering
\subfigure[]{
\includegraphics[height=3.1cm]{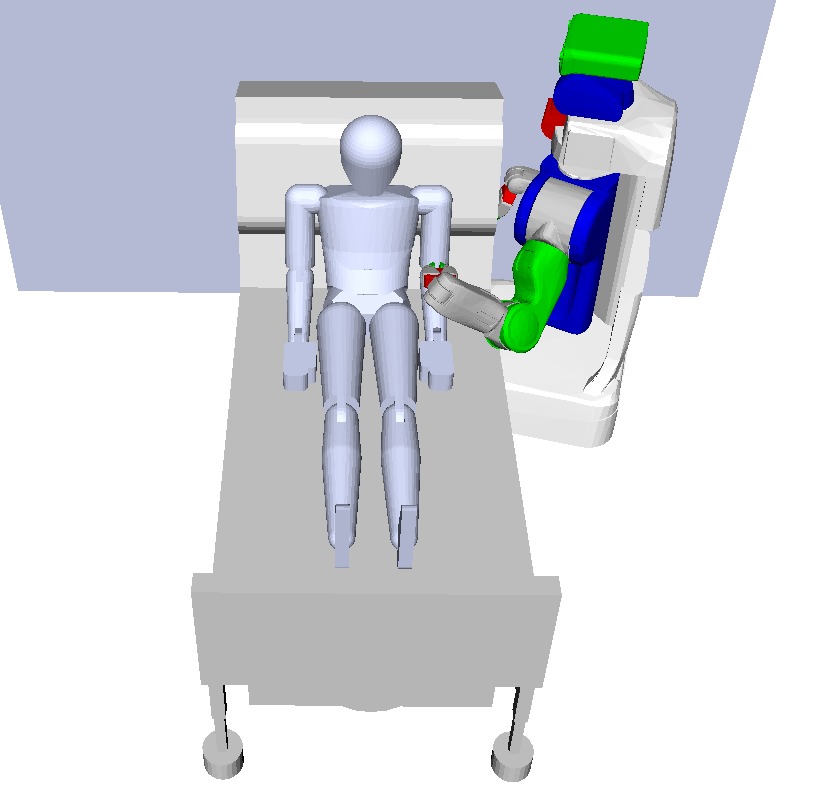}}
\hfill
\subfigure[]{
\includegraphics[height=3.1cm]{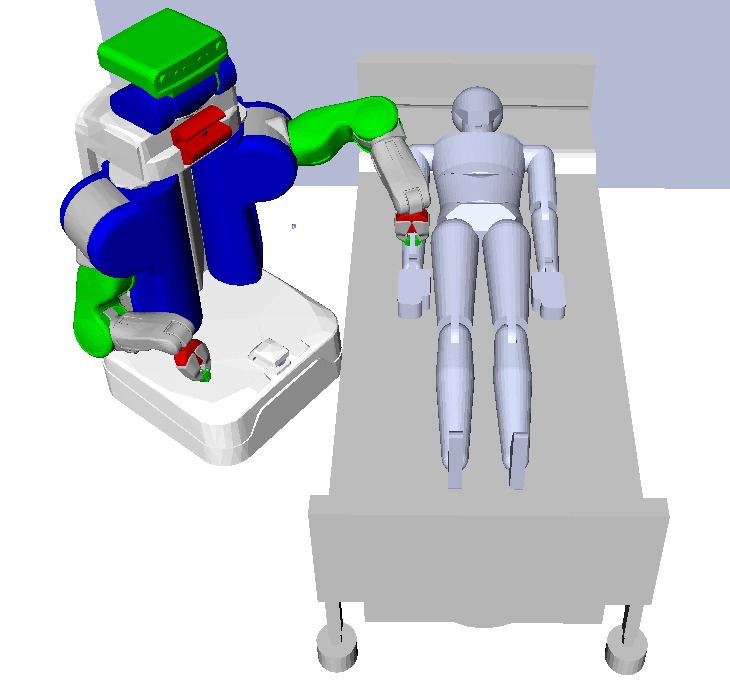}}
\hfill
\subfigure[]{
\includegraphics[height=3.1cm]{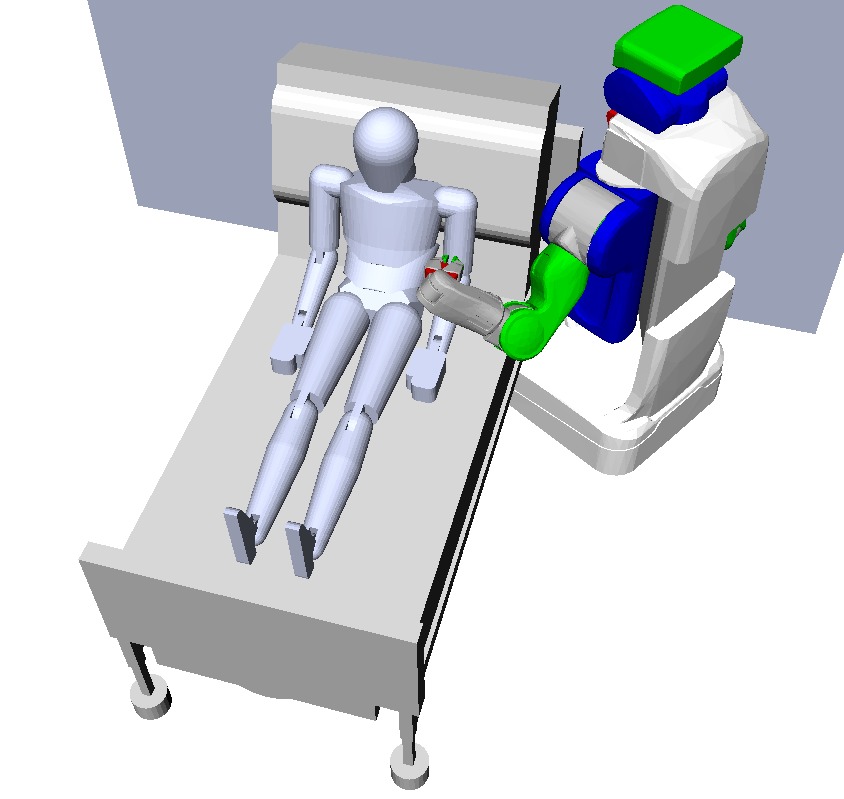}}
\hfill
\subfigure[]{
\includegraphics[height=3.1cm]{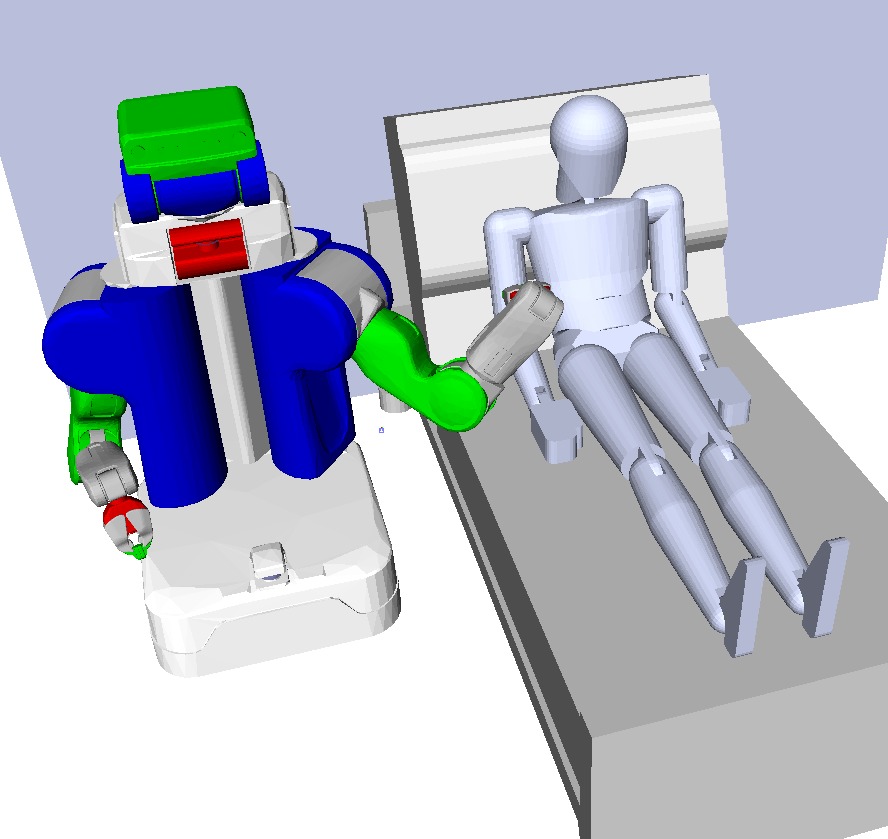}}
\hfill
\subfigure[]{
\includegraphics[height=3.1cm]{legs_bed_1.jpg}}\\
\subfigure[]{
\includegraphics[height=3.1cm]{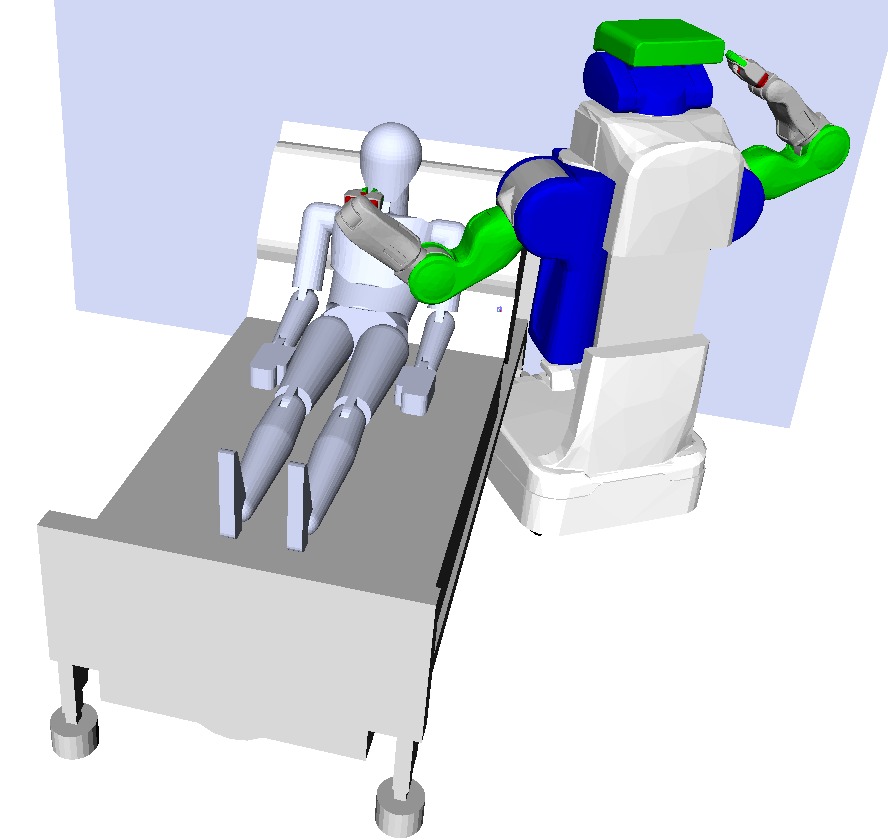}}
\hfill
\subfigure[]{
\includegraphics[height=3.1cm]{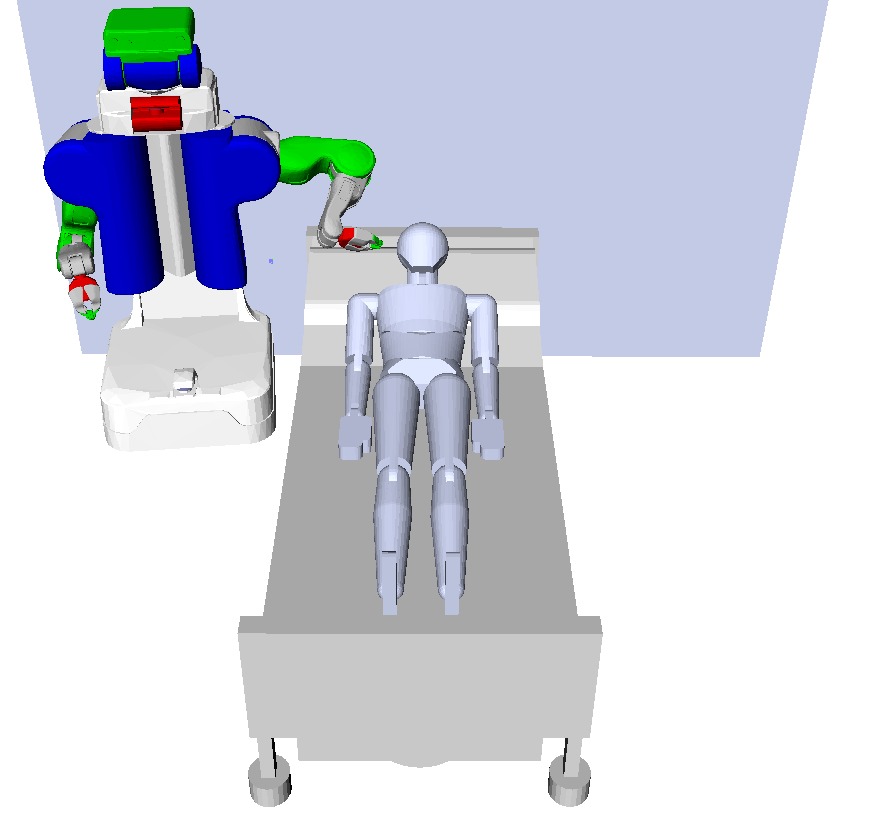}}
\hfill
\subfigure[]{
\includegraphics[height=3.1cm]{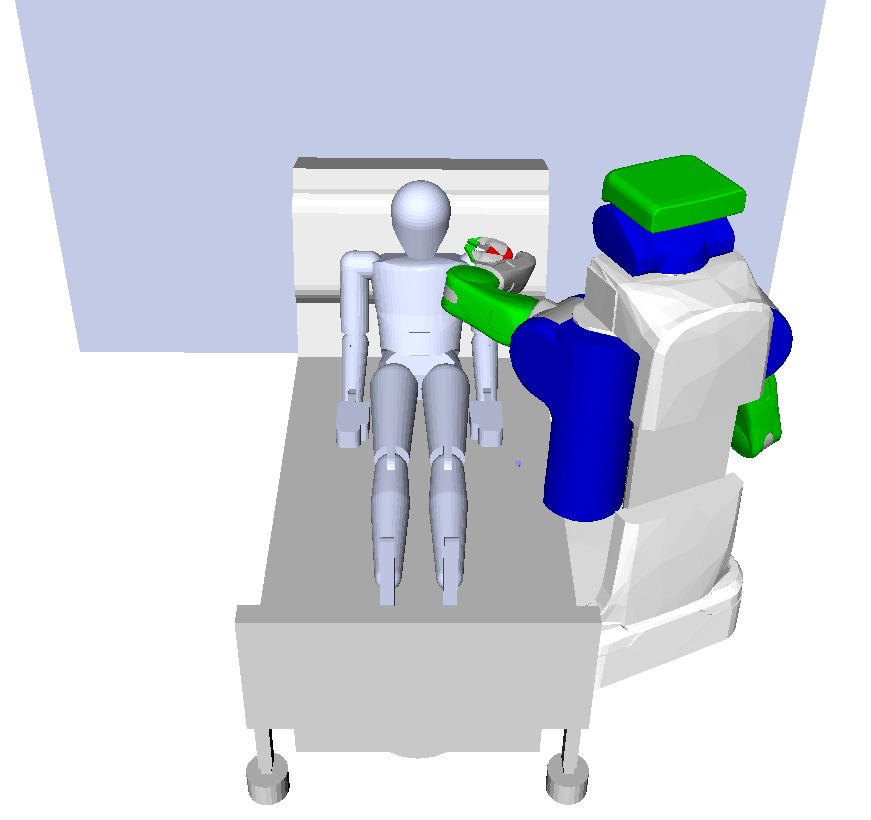}}
\hfill
\subfigure[]{
\includegraphics[height=3.1cm]{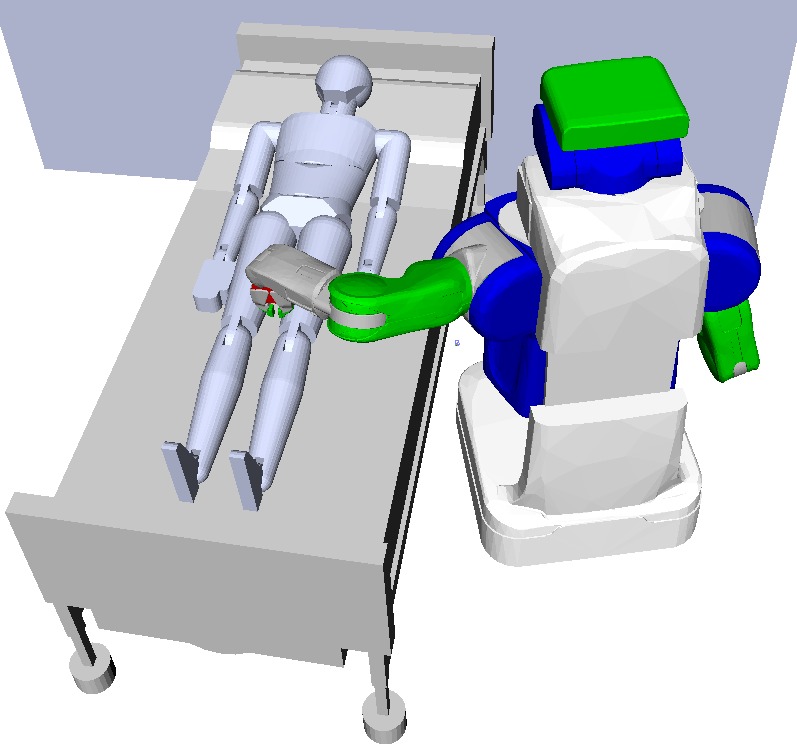}}
\hfill
\subfigure[]{
\includegraphics[height=3.1cm]{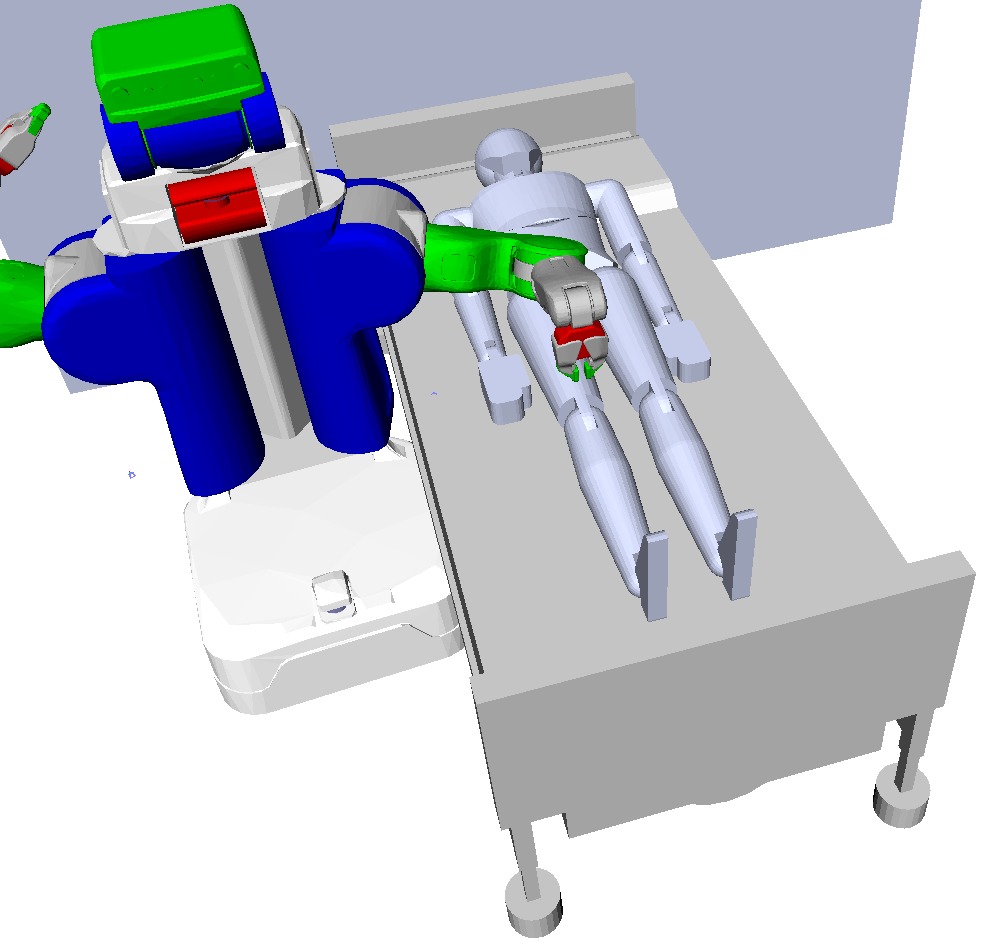}}\\
\subfigure[]{
\includegraphics[height=3.1cm]{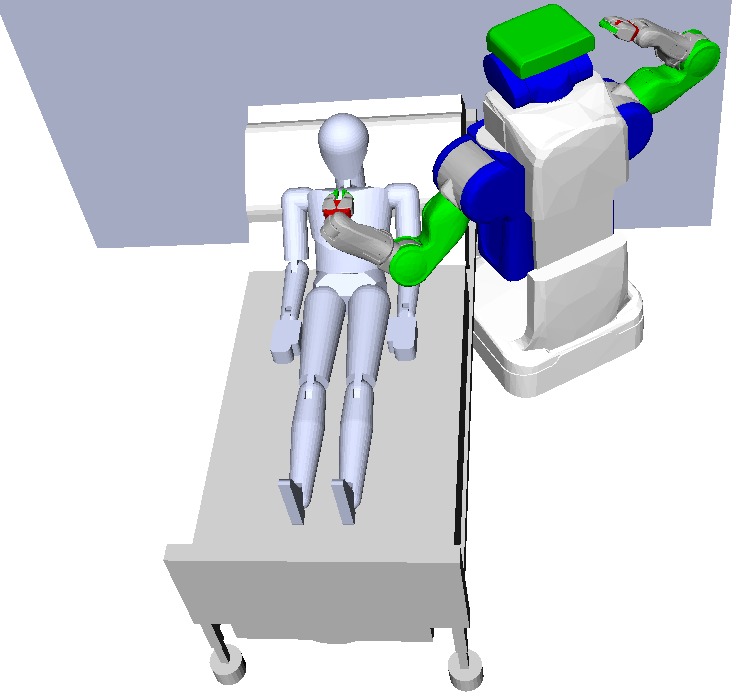}}
\hfill
\subfigure[]{
\includegraphics[height=3.1cm]{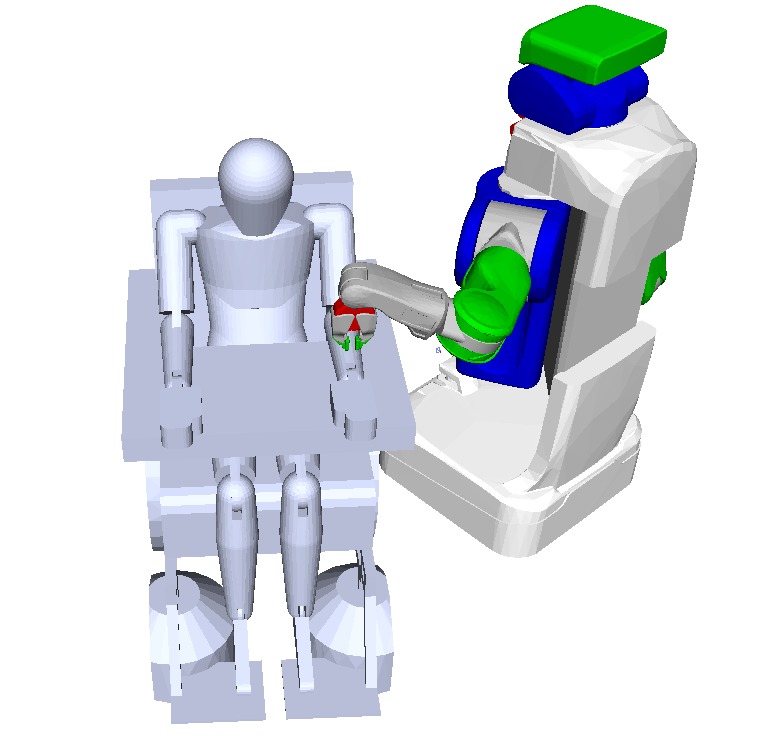}}
\hfill
\subfigure[]{
\includegraphics[height=3.1cm]{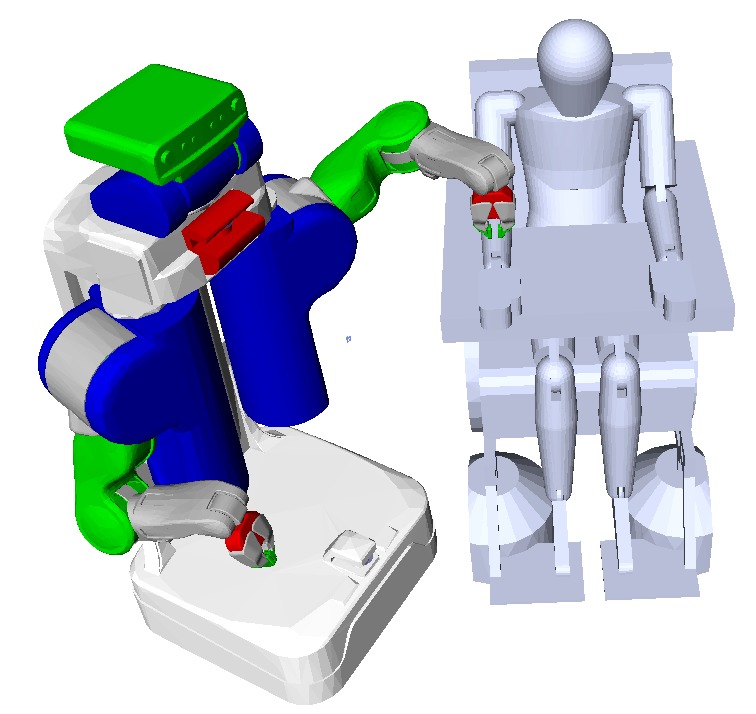}}
\hfill
\subfigure[]{
\includegraphics[height=3.1cm]{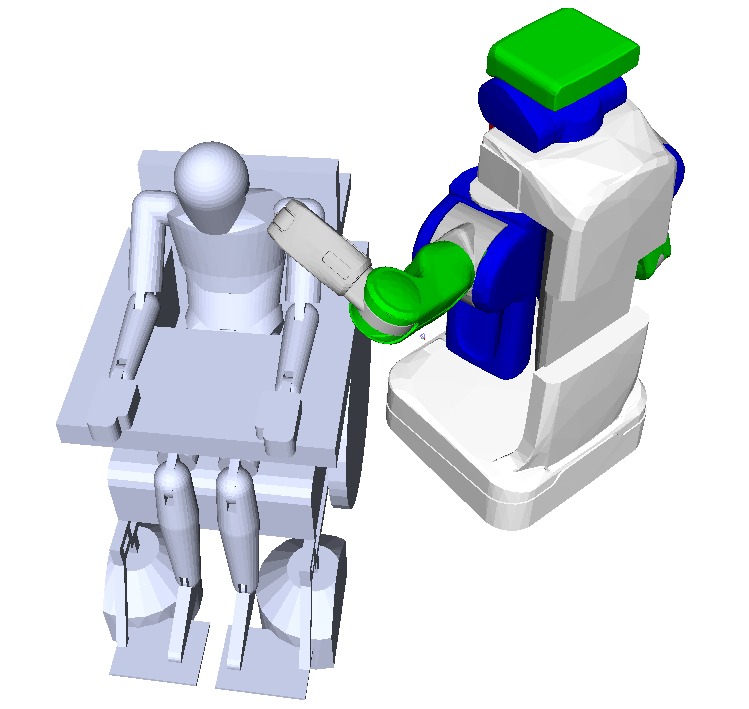}}
\hfill
\subfigure[]{
\includegraphics[height=3.1cm]{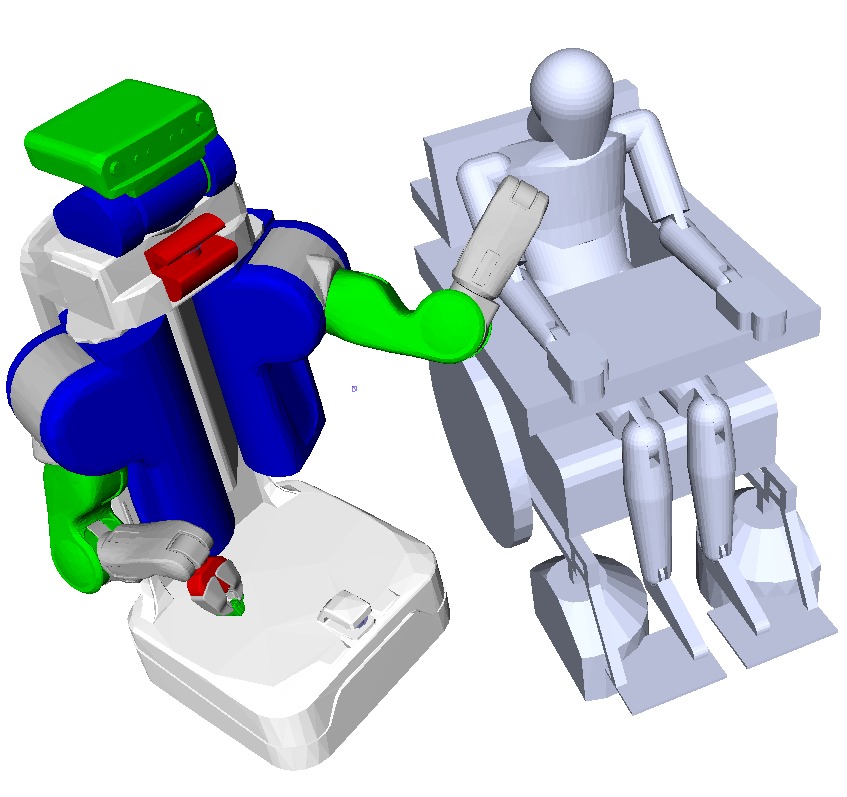}}\\
\subfigure[]{
\includegraphics[height=3.1cm]{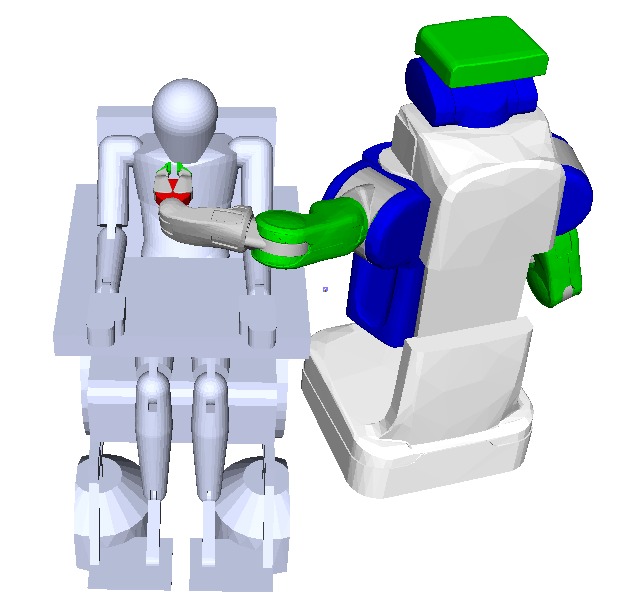}}
\hfill
\subfigure[]{
\includegraphics[height=3.1cm]{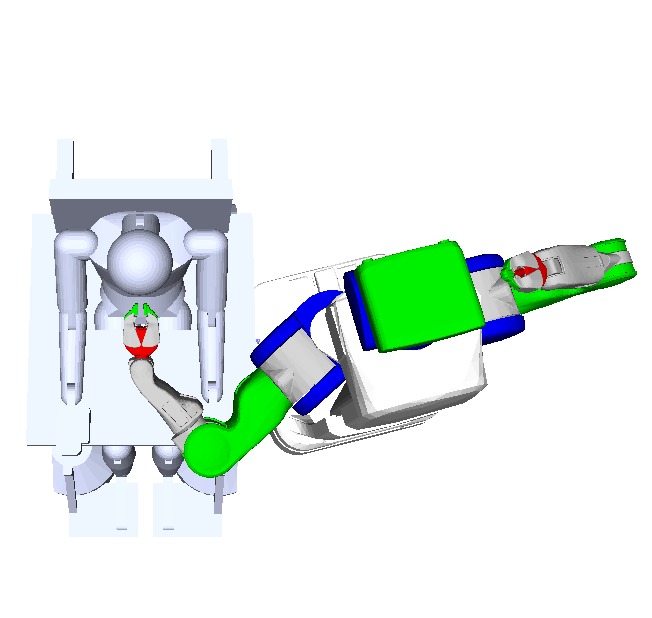}}
\hfill
\subfigure[]{
\includegraphics[height=3.1cm]{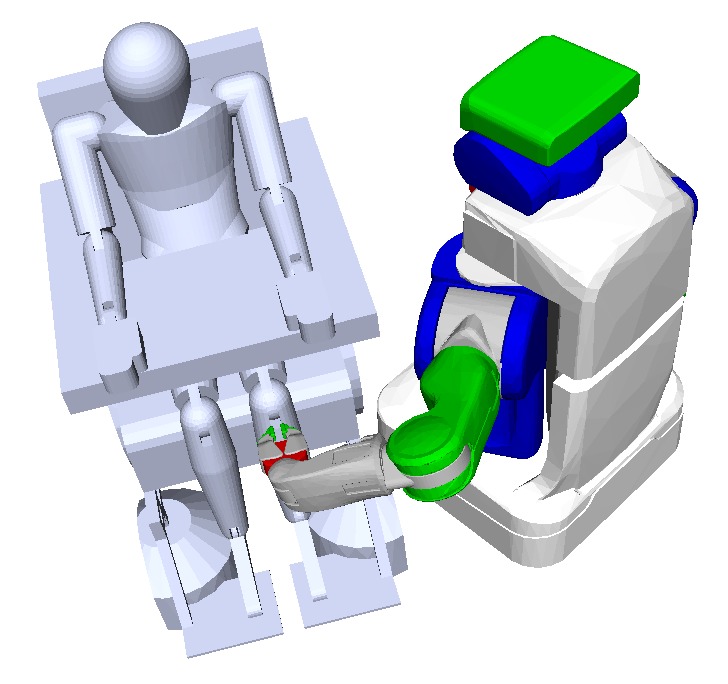}}
\hfill
\subfigure[]{
\includegraphics[height=3.1cm]{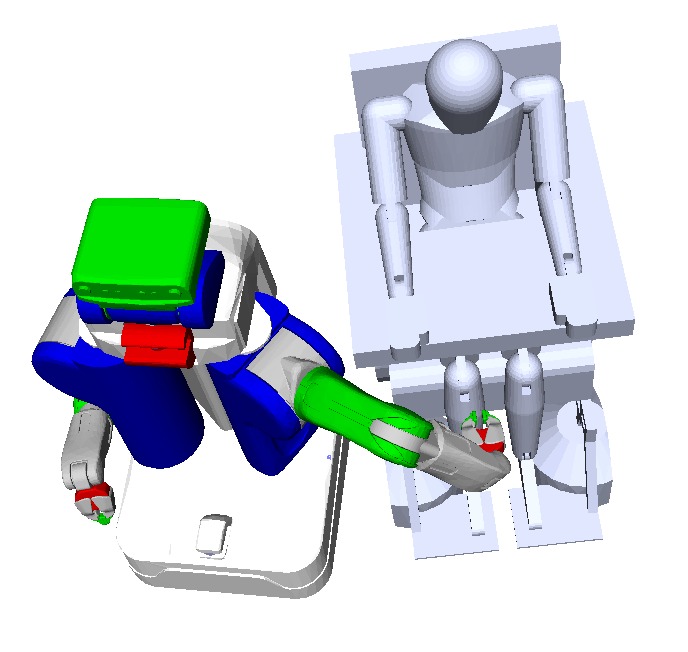}}
\subfigure[]{
\includegraphics[height=3.1cm]{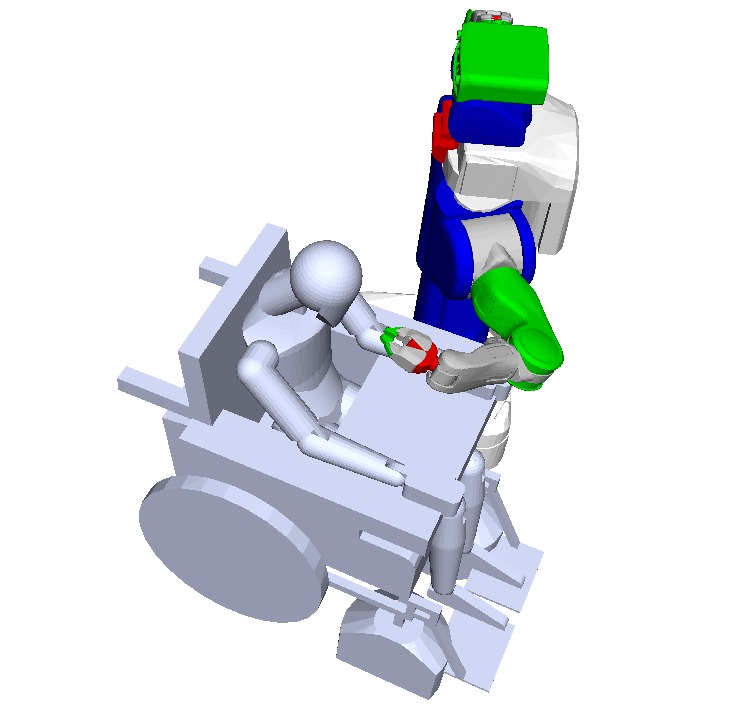}}
\caption{Visualization of the robot configurations selected by TOC for each task. Two images are used when TOC selected two configurations for a task task. 
The images show for the robotic bed environment: (a)~cleaning arms config~\#1 
(b)~cleaning arms config~\#2 (c)~scratching left upper arm 
(d)~scratching right upper arm (e)~cleaning legs (f)~wiping mouth
(g)~shaving config~\#1 (h)~shaving config~\#2 (i)~scratching left knee 
(j)~scratching right knee (k)~feeding.
The images show for the wheelchair environment:
(l)~arm cleaning config~\#1 
(m)~arm  bed config~\#2 (n)~scratching left upper arm 
(o)~scratching right upper arm (p)~wiping mouth
(q)~shaving (r)~scratching left knee 
(s)~scratching right knee (t)~feeding.
}
\label{fig:configuration_visualization}
\end{figure*}

\Urlmuskip=0mu plus 3mu
We ran all simulations in OpenRAVE (\url{http://www.openrave.org/} from \citet{diankov2008openrave}), for which we created environment models with a PR2 robot and a 
model of an average male human placed either in a wheelchair or in a robotic bed. 
The human model dimensions come from \citet{human_measurement}. 
The PR2 is a mobile manipulator
made by Willow Garage with two 7-DoF arms.
\Urlmuskip=0mu plus 0mu
The models we created for the robotic bed and the wheelchair match Autobed, a modified Invacare 
5401IVC full electric hospital bed \citep{grice2016resna} and a Sunrise Medical Quickie 2 wheelchair with 
overlap table, respectively. The casters on the bed and wheelchair are represented by swept volumes. For the wheelchair, we removed the part of the casters' swept volumes that extends to the sides of the chair to increase free space around the chair. 
We assume that the user would ensure that the casters are not pointing out from the chair.

Figure~\ref{fig:configuration_visualization} shows the configurations selected by our implementation of TOC for each task, given the observation, $\hat{h}$, that the person was positioned in the center of the bed or wheelchair.

The robotic bed environment has the bed in front of a wall,
to emulate how beds are often positioned in rooms. The robotic bed can raise up to \SI{25}{\centi\metre} and can increase the angle of its head rest up to \SI{75}{\degree}. For the wheelchair environment we
gave the human the ability to rotate its neck up to \SI{45}{\degree} in either direction about the Z-axis.
Figure~\ref{fig:shaving_chair} shows the neck rotated \SI{45}{\degree}. These two environments (robotic bed and wheelchair) were selected to demonstrate many of the functionalities of the TOC framework. The robotic bed environment demonstrates how TOC can select configurations for multiple robots in the environment, how TOC can handle controllable and uncontrollable parameters of the environment, and how it handles uncontrollable and environment-driven parameters of the user. TOC treats additional robots the same as controllable objects, adding their parameters to the robot configuration, as it does with the robotic bed's parameters. TOC considers the position of the person on the bed as an uncontrollable parameter and considers the other parameters of the user (the configuration of the person's joints) as environment-driven parameters. The wheelchair environment demonstrates how TOC can make use of free parameters in the user model. TOC considers the rotation of the person's neck as a free parameter.



\begin{figure*}[t]
\centering
\includegraphics[width=\textwidth]{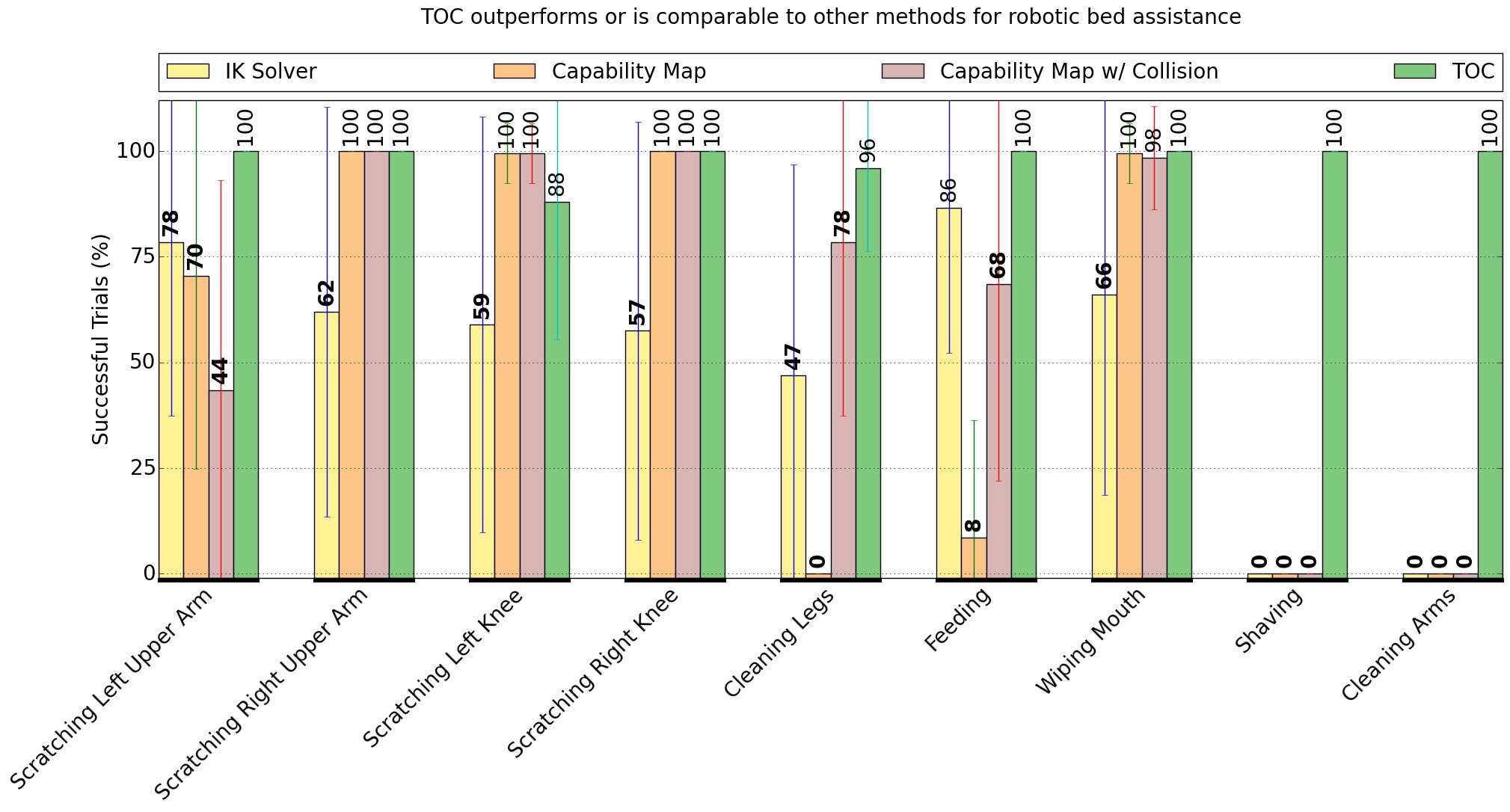}\\
\caption{Comparison of performance between TOC and three baseline methods averaged over 200 Monte-Carlo simulations of state estimation error for tasks in the robotic bed environment. Bold numbers have statistically significant ($p<0.01$) difference from the TOC result in a Wilcoxon Rank-Sum test. Error bars show one standard deviation. TOC chose to use a single configuration for all tasks other than the shaving and cleaning arms tasks in this environment. Baseline methods could only select a single configuration.}

\label{fig:baseline_comparison_results_autobed}
\end{figure*}

\begin{figure*}[t]
\centering
\includegraphics[width=\textwidth]{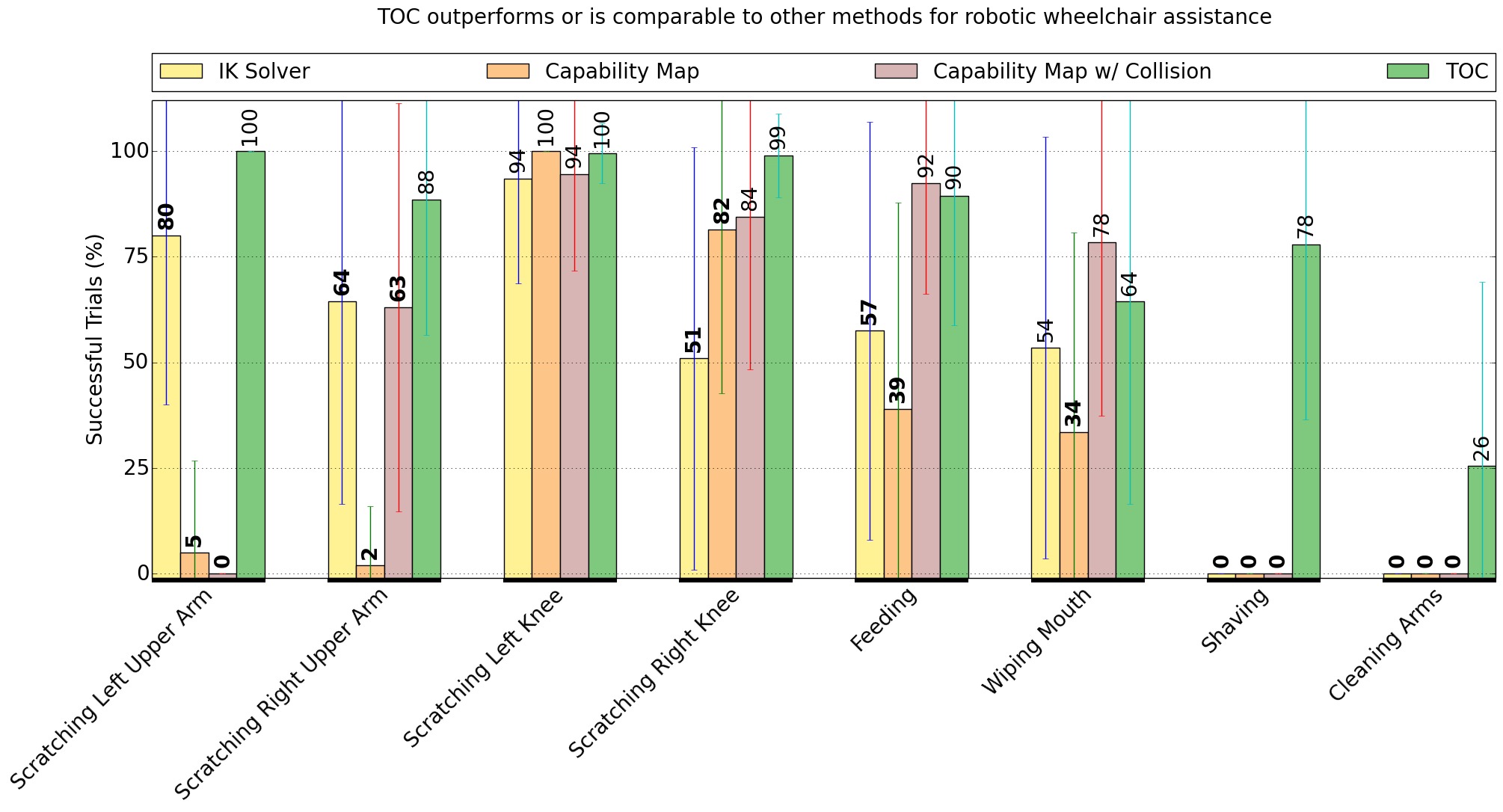}
\caption{Comparison of performance between TOC and three baseline methods averaged over 200 Monte-Carlo simulations of state estimation error for tasks in the wheelchair environment. Bold numbers have statistically significant ($p<0.01$) difference from the TOC result in a Wilcoxon Rank-Sum test. Error bars show one standard deviation. TOC chose to use a single configuration for all tasks other than the cleaning arms task in this environment. Baseline methods could only select a single configuration.}
\label{fig:baseline_comparison_results_chair}
\end{figure*}

\subsection{Evaluation Against Baselines}\label{ssec:evaluation_vs_baseline}
We compared the performance of TOC against three baseline methods in Monte Carlo simulations of Gaussian error introduced in the person's position (e.g., translating around on the bed while the bed remains stationary) and in the robot's base pose (e.g., translating and rotating the robot from the selected robot configuration). Each method estimated an optimal set of robot configurations, $\boldsymbol{\hat{r}}^{*}$, given the observation, $\hat{h}$, that the person was positioned in the center of the bed or wheelchair. Because the goals are sparse and represent a more complicated task, we considered a trial successful if, from the robot configurations selected by the method, the PR2 could reach all goal poses despite the error introduced in the Monte Carlo simulation. Otherwise, the trial was a failure.
We performed this evaluation for all 9 modeled tasks in both environments except the cleaning legs tasks for the wheelchair environment because the overlap table blocks access to the thighs, preventing successful performance of the task.

All introduced error was normally distributed around 0. For the robotic bed environment, the standard deviation for the human's pose was \SI{2.5}{\centi\metre} translation in the global X direction and \SI{5.0}{\centi\metre} translation in the global Y direction. Rotations of the human in bed were not considered because small rotations about the head results in large movements of the legs. For the wheelchair environment, the standard deviation for the human's pose was \SI{2.5}{\centi\metre} translation in the global X direction, \SI{5.0}{\centi\metre} translation in the global Y direction, and \SI{5}{\degree} rotation about the human head's Z axis. The standard deviation for the PR2's position was \SI{1.0}{\centi\metre} in the global X and Y directions and \SI{5}{\degree} rotation about the robot's Z axis. We selected these error distributions from typical error in human pose estimation and PR2 servoing in our previous work \citep{kapusta2016collaboration}. As described in Sections~\ref{ssec:environment_modeling} and~\ref{ssec:user_modeling}, models used for selecting configurations had a small safety margin of $\sim$\SI{3}{\centi\metre}. Models used for testing had no safety margin.

For fair comparison, all methods in this evaluation were given matching seeds for their optimization via CMA-ES as well as for their error in Monte-Carlo simulation. For the CMA-ES optimization, all methods were given a population size of 40, a maximum number of iterations of 1000, and the opportunity to restart with double the population if the optimization ran out of iterations before converging within some tolerance. All methods also had the same heuristics for driving the search towards configurations that may have collision-free IK solutions.
We assigned appropriate bounds on parameters based on the environment (e.g., slightly beyond reach of the bed). We initialized the parameters to aid coverage in the search, giving two initial locations, one position on one side of the bed or wheelchair and one on the other side. Baseline methods were allowed two searches, one for each initialization, and selected the single best configuration. TOC jointly optimized its two configurations from their respective initialization locations. Thus, all methods were given the comparable initializations and bounds. 

\begin{figure*}[t]
\includegraphics[width=\textwidth]{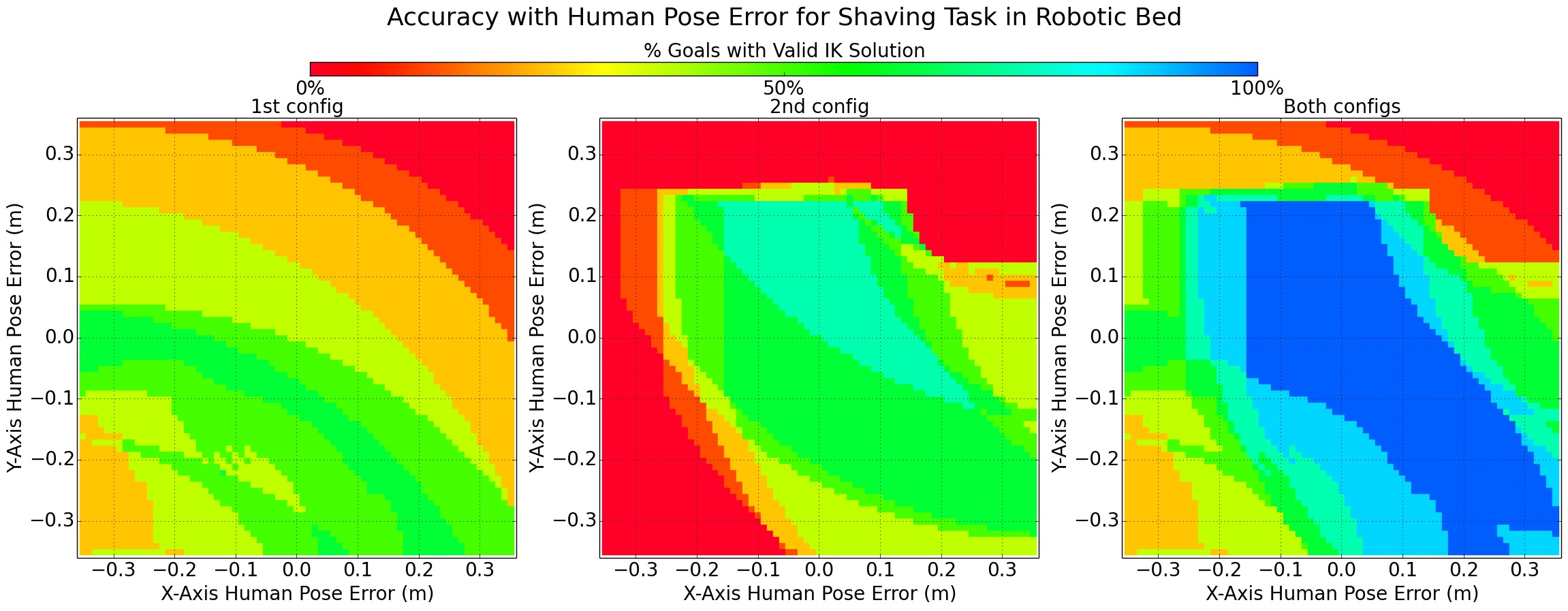}\\
\vspace{0.1cm}
\hspace{0.55cm}
\includegraphics[height=5cm]{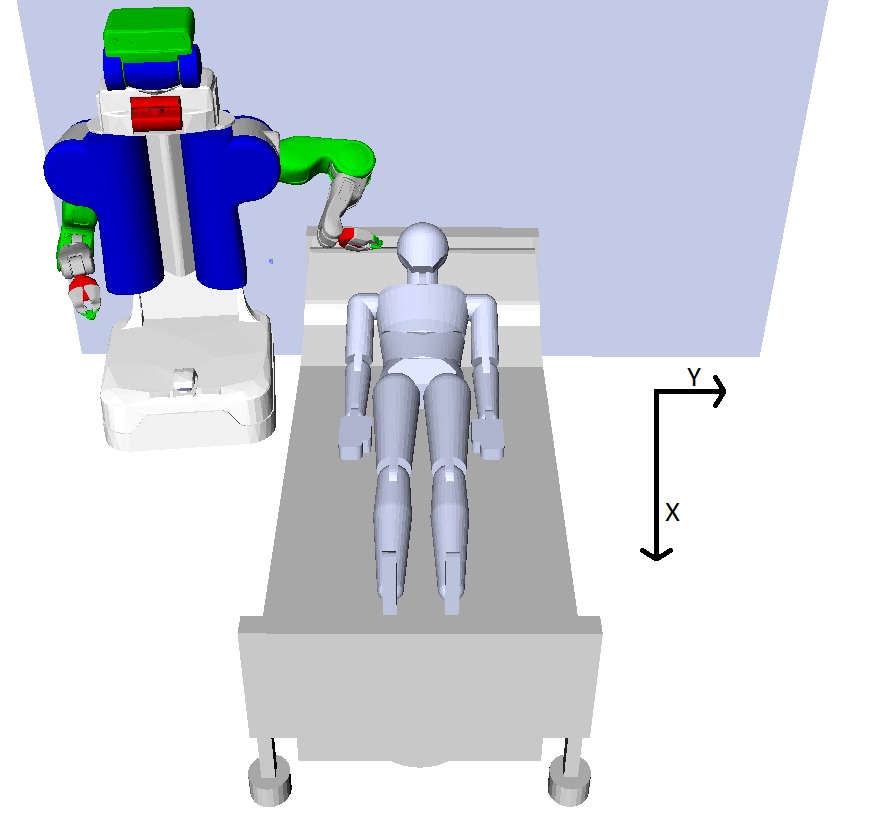}
\hspace{0.25cm}
\includegraphics[height=5cm]{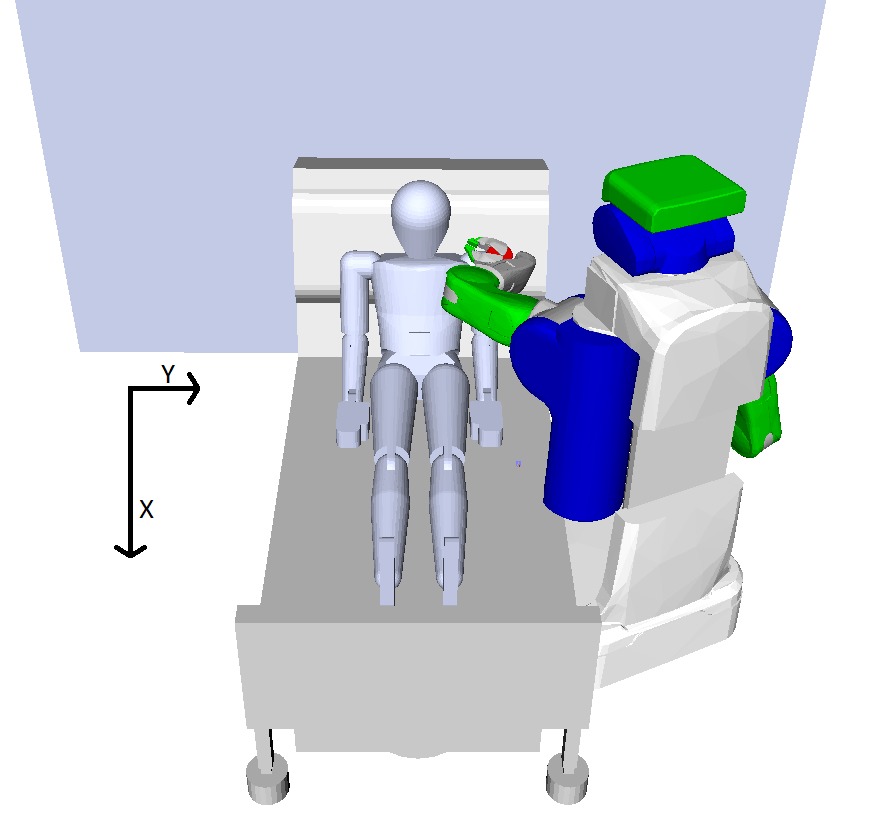}
\hspace{0.25cm}
\raisebox{1.25cm}{\includegraphics[height=2.5cm]{shaving_bed_1_axes.jpg}}
\raisebox{1.25cm}{\includegraphics[height=2.5cm]{shaving_bed_2_axes.jpg}}

\caption{Visualization of the robustness of TOC's selected configurations for the shaving task in the robotic bed. (Top) Percentage of goals reached from the first, second, and both configurations for error in 1cm increments in the x-y position of the person. (bottom) The first and second configurations of the PR2, and the two configurations combined on the right.  Color is necessary to interpret this figure. The blue region represents when all goals can be reached.
}\label{fig:robustness_shaving_autobed}
\end{figure*}

\begin{figure*}[t]
\includegraphics[width=\textwidth]{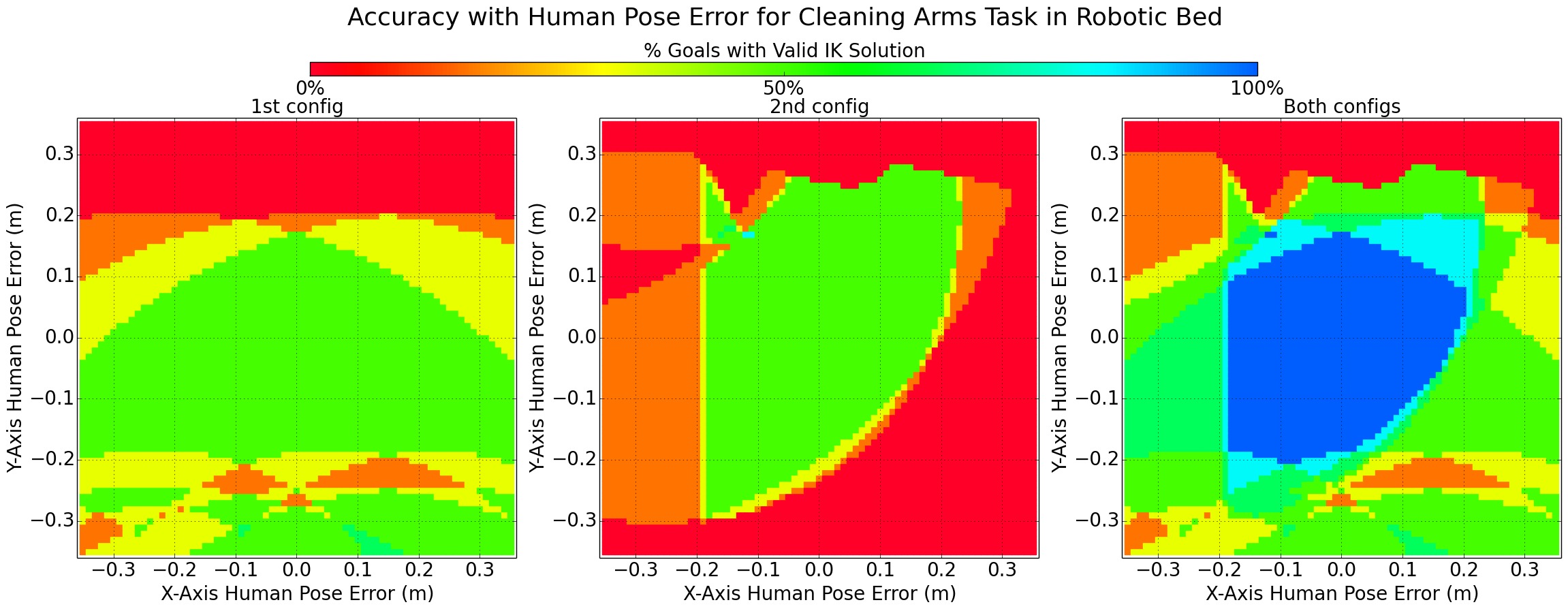}\\%
\vspace{0.1cm}
\hspace{0.5cm}
\includegraphics[height=5cm]{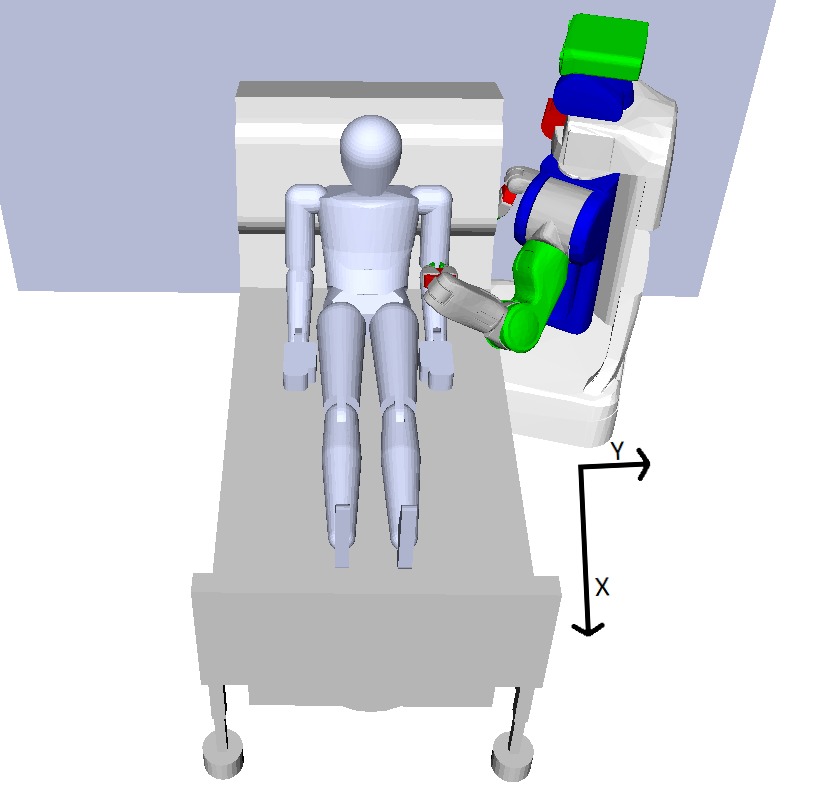}
\hspace{0.1cm}
\includegraphics[height=5cm]{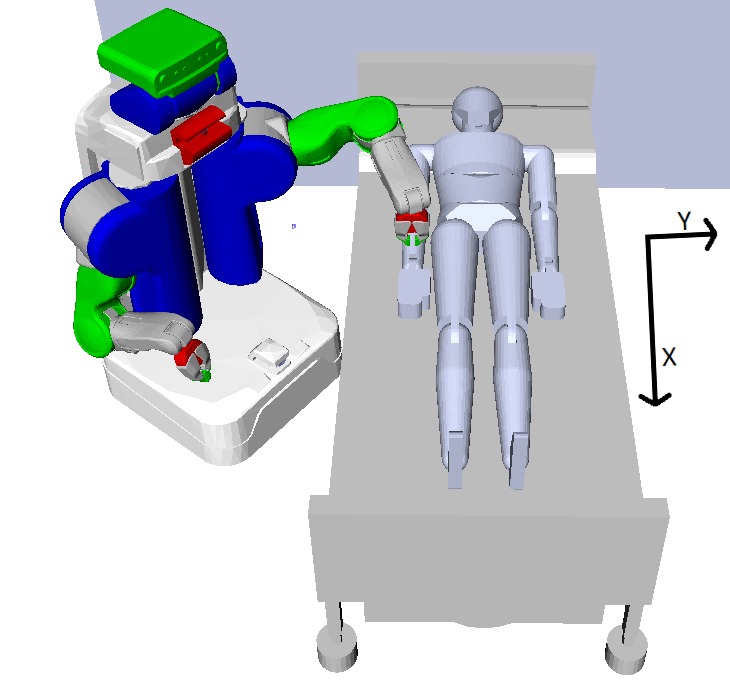}
\hspace{0.4cm}
\raisebox{1.25cm}{\includegraphics[height=2.5cm]{arms_bed_1_axes.jpg}}
\raisebox{1.25cm}{\includegraphics[height=2.5cm]{arms_bed_2_axes.jpg}}

\caption{Visualization of the robustness of TOC's selected configurations for the arm cleaning task in the robotic bed. (Top) Percentage of goals reached from the first, second, and both configurations for error in 1cm increments in the x-y position of the person. (bottom) The first and second configurations of the PR2, and the two configurations combined on the right.  Color is necessary to interpret this figure. The blue region represents when all goals can be reached. }\label{fig:robustness_arms_autobed}
\end{figure*}

 \begin{figure*}[t]
\includegraphics[width=\textwidth]{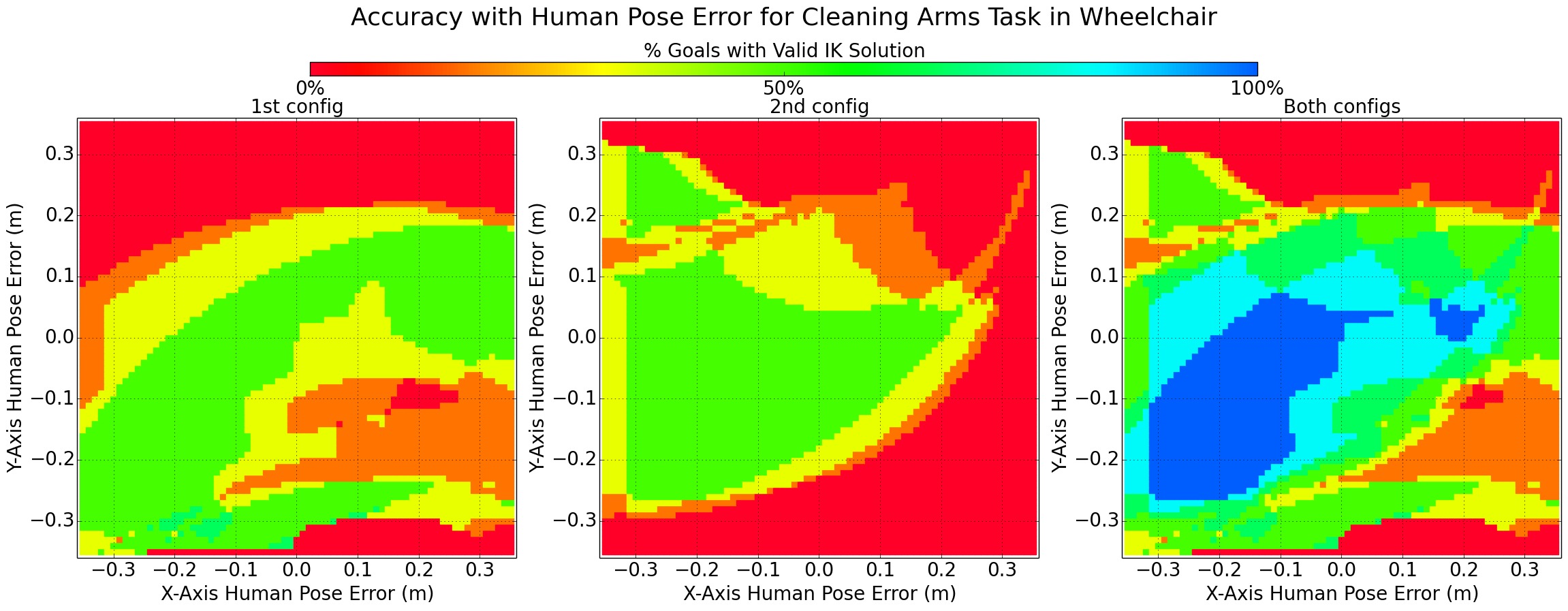}\\%
\vspace{0.1cm}
\hspace{0.6cm}
\includegraphics[height=5cm]{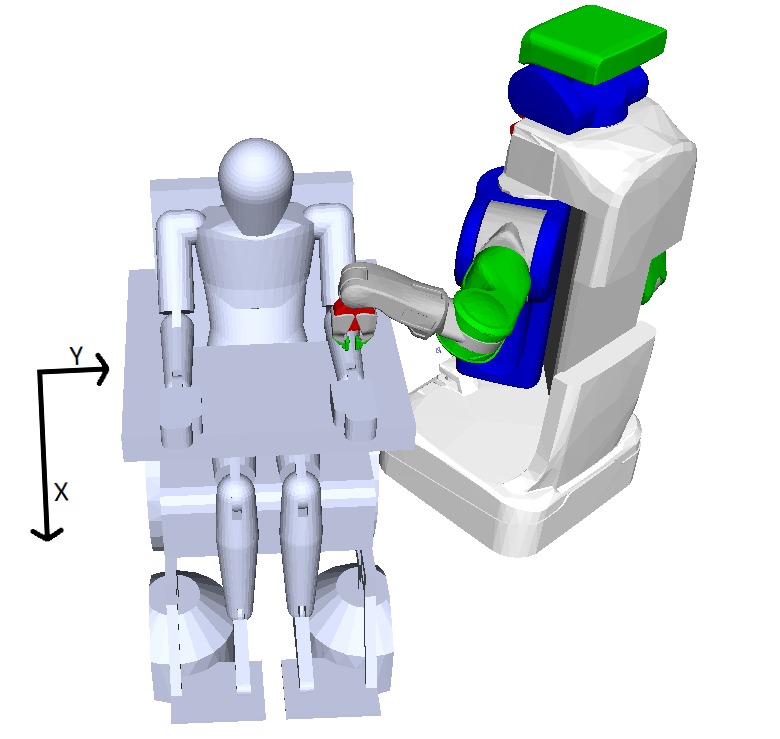}
\hspace{0.4cm}
\includegraphics[height=5cm]{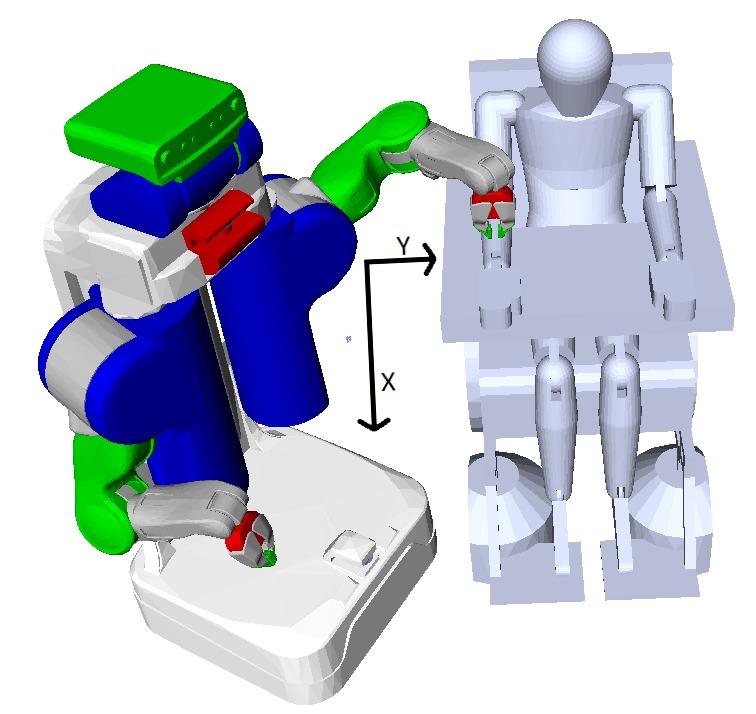}
\hspace{0.1cm}
\raisebox{1.25cm}{\includegraphics[height=2.5cm]{arms_chair_1_axes.jpg}}
\raisebox{1.25cm}{\includegraphics[height=2.5cm]{arms_chair_2_axes.jpg}}

\caption{Visualization of the robustness of TOC's selected configurations for the arm cleaning task in the wheelchair. (Top) Percentage of goals reached from the first, second, and both configurations for error in 1cm increments in the x-y position of the person. (bottom) The first and second configurations of the PR2, and the two configurations combined on the right.  Color is necessary to interpret this figure. The blue region represents when all goals can be reached. }\label{fig:robustness_arms_chair}
\end{figure*}

\subsubsection{Baselines}
We implemented three baselines from literature to compare against TOC, one based on IK and two based on the robot capability map. An overview of these and other related methods from literature can be found in Section~\ref{sec:related_work}. These methods selected a single configuration for both the PR2 and the robotic bed and made use of the human's free parameter (neck rotation) in the wheelchair environment.

\paragraph{Inverse-Kinematics (IK) Solver-based Baseline}
IK solver-based methods to select a robot base pose for a task are common in literature. The method we implemented uses CMA-ES to search for a robot configuration where the robot has a collision-free IK solution to all goal end effector poses. We used the ikfast module within the OpenRAVE simulation environment to determine if a collision-free IK solution existed for each robot configuration. 

\paragraph{Capability Map-based Baselines}
Various methods from literature use the capability map \citet{zacharias2007capturing}. We implemented two baseline methods  roughly based on \citet{zacharias2009using}. For these methods we first created a capability map using OpenRAVE's kinematic reachability module. To create the capability map, the module discretized 3D space around the robot's arm into 3D points and discretized the range of possible orientations around each 3D point. The capability score (also known as reachability score) for each point is the percentage of orientations for which the robot has a valid IK solution. These scores are calculated offline and saved. These two methods use CMA-ES to search for a robot configuration that maximizes the average capability score for all goal poses. The score of a goal pose is the score of the closest 3D point from the capability map. The first capability map-based baseline considered capability scores without regard to the environment. The second  gave goal poses a 0 score if a collision-free IK solution could not be found to that pose in the environment. 

\subsubsection{Results}
The results for each task for the robotic bed and wheelchair are shown in Figures~\ref{fig:baseline_comparison_results_autobed} and~\ref{fig:baseline_comparison_results_chair}, respectively. TOC's average success rate was higher than or comparable to baseline methods in all tasks. Statistically significant difference from the TOC result ($p<0.01$ in a Wilcoxon Rank-Sum test) is indicated in the figures with bold numbers.

TOC had an overall average success rate of 90.6\%, compared to 50.4\% for IK, 43.5\% for capability map, and 58.9\% for capability map with collision checking.
The overall differences between baseline results and TOC are statistically significant ($p<0.01$) in a Wilcoxon Rank-Sum test. 
TOC chose to use a single configuration for all tasks other than the shaving and cleaning arms tasks in this environment in the bed environment and it used a single configuration for all but the cleaning arms task in the wheelchair environment. TOC achieved higher average success rates both for tasks for which it used one and two configurations. 
This result suggests that benefit from TOC comes from more than just from using 2 configurations over one.
For tasks that require more than one robot configuration, baseline methods failed; they could only select a single configuration. TOC jointly optimizes 2 configurations, allowing it to succeed in these challenging tasks. 

\subsection{Quantifying Robustness}\label{ssec:evaluation_robustness}
In Figures~\ref{fig:robustness_shaving_autobed}, \ref{fig:robustness_arms_autobed} and~\ref{fig:robustness_arms_chair} We visualize the robustness of robot configurations
selected by TOC for the shaving and cleaning arms tasks in the two environments. These figures show the percentage of goal poses that have collision-free IK solutions (indicated by the color) for varying error in the 
person's position on the bed or wheelchair (the X-Y axes). Notable in these figures is the success region in blue, where all goals
are reachable, as well as how the two configurations combine to reach all goals. For pose estimation error in the success region, 
the PR2 would still be able to successfully perform the task. 
TOC opted to use two configurations for each of the tasks shown. The success region is large and surrounds the origin for shaving and cleaning arms in bed, which is why 100\% of the trials were successful for these tasks in Figure~\ref{fig:baseline_comparison_results_autobed}. The success region is less centered around the origin for the cleaning arms task in the wheelchair, hence its lower percentage of successful trials in Figure~\ref{fig:baseline_comparison_results_chair}. The Monte Carlo simulations randomly sampled in these, as well as other, degrees of freedom and sampling outside the success region results in a failed trial. These figures suggests that the task may be easier for the robot to perform for a person in bed. Closer observation of the task shows that, because the wheelchair is tall, the goals poses for the cleaning arms task are vertically higher in the PR2's workspace and the arm have relatively low JLWKI when reaching those goals. 

\begin{figure}[h]
\centering
\includegraphics[width=\columnwidth]{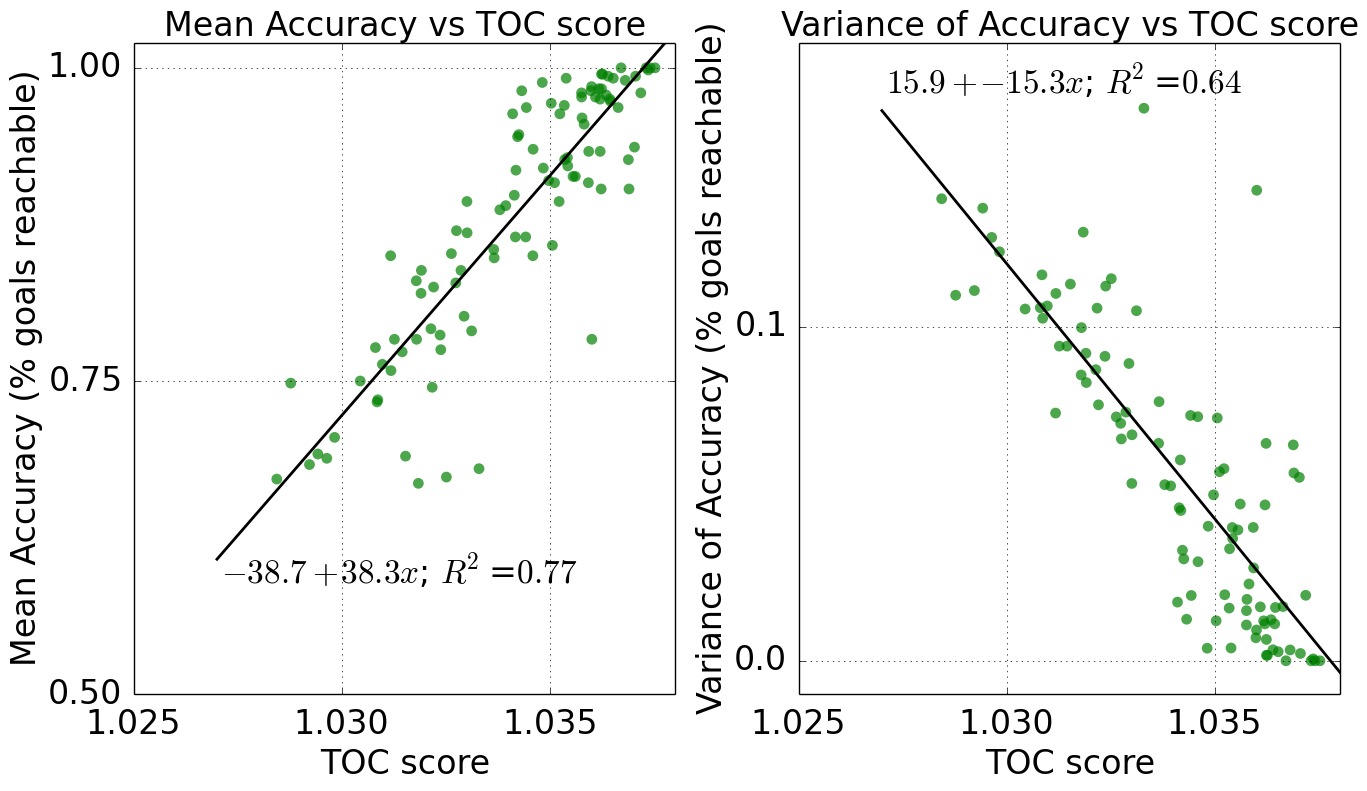}
\caption{Increasing TOC score is correlated to accuracy and inversely correlated to variance in accuracy. TOC score above 1.0 means a collision-free IK solution exists to all goals. The amount above 1.0 is the weighted TC-manipulability score for the configuration. }
\label{fig:evaluation_scoring}
\end{figure}

\subsection{Evaluation of TOC Objective Function}\label{ssec:evaluation_scoring}
TOC searches for a set of robot configurations that maximizes its objective function, which we will 
call its score for simplicity. The assumption therein is that higher values of the score are correlated 
with better robot configurations, that are more robust to error. To test this assumption, we evaluated the relationship 
between the TOC score and the accuracy (the percentage of goals that are reachable) for robot configurations in the same evaluation described in Section~\ref{ssec:evaluation_vs_baseline}. Note that we chose to compare against accuracy in this evaluation because it can convey more information than success, which is binary. A similar correlation can be seen for success. For the wiping mouth task in the robotic bed, we sampled robot configurations with TC-reachability of 1 (i.e., all goal poses have collision-free IK solutions) and compared their mean and variance in accuracy over 200 Monte Carlo simulations with their TOC score. Figure~\ref{fig:evaluation_scoring} shows the results of the analysis. Higher TOC score is correlated with accuracy and inversely correlated with variance in accuracy.

\section{Discussion}\label{sec:discussion}
We have shown results with a single type and source of error. However, we expect TOC to be able to deal with many types and sources of error. If we knew all sources of error apriori, we could explicitly model them, to improve selection of robot configurations. 
Such a method would be similar to those from \citet{hsu1999placing} and \citet{stulp2009learning}, described in Section~\ref{ssec:rw_robot_configurations}. It should be noted, however, that movement of the person (e.g., pose estimation error) does not directly translate into motion of the robot's base, so directly using these evaluations as part of a method to select the robot's configuration can be difficult. Our results provide evidence that TOC can often select robot configurations that are more robust to error in the person's pose than baseline methods. This is seen in many of the tasks and environments we examined, but is most pronounced in more challenging tasks that may need more than a single robot configuration to complete. 

We observed that baseline methods work well for many tasks and they selected configurations, on average, faster than TOC. For its offline computations, TOC took on average 70 minutes for each task running in a single thread on a a 64-bit, 14.04 Ubuntu operating
system with 8 GB of RAM and a 3.40 GHz Intel Core
i7-3770 CPU. The IK solver baseline we implemented runs faster, on average 6 minutes and often less than 1 minute for each task. These results suggest that this baseline may work well for many tasks that only require a single robot configuration and where there is little error between model and reality. The capability map with collision detection also performed well on many tasks and took on average 34 minutes to select a configuration. Note that this is the time to run these algorithms either offline within our framework or online. As in assistive robotics, in many robotics applications the speed of online selection is important and 6 minutes, 34 minutes, and 70 minutes are all too long for a user to wait for a robot to decide where to move. Additionally, there may be specifically important or common tasks that take place in environments that are known beforehand. In this case, it can be valuable to take additional time offline to find solutions, using our framework, to take advantage of prior knowledge and speed up the responsiveness of the online process.

We sought to handle fairly the comparison between TOC and baseline methods by using the same sampling for each. However, because of the nature of the search problem, using other search methods, additional heuristics, or other meta-parameters for the search (e.g., different  population size) may find better configurations than those found in our evaluations. 

We observed a trade-off when selecting robot configurations. In general, moving the PR2 closer tends to improve the arm's dexterity, but tends to make collisions more likely. An assumption in this work is that contact is bad and should be avoided. However, \citet{grice2013whole} found that contact can be both beneficial and acceptable during robotic assistance. Allowing contact can increase the space of reachable poses, and there are methods for controlling contact safely \citep{killpack2016model}.


\subsection{Design Application}\label{ssec:discussion_design}
\begin{figure}[h]
\centering
\includegraphics[width=\columnwidth]{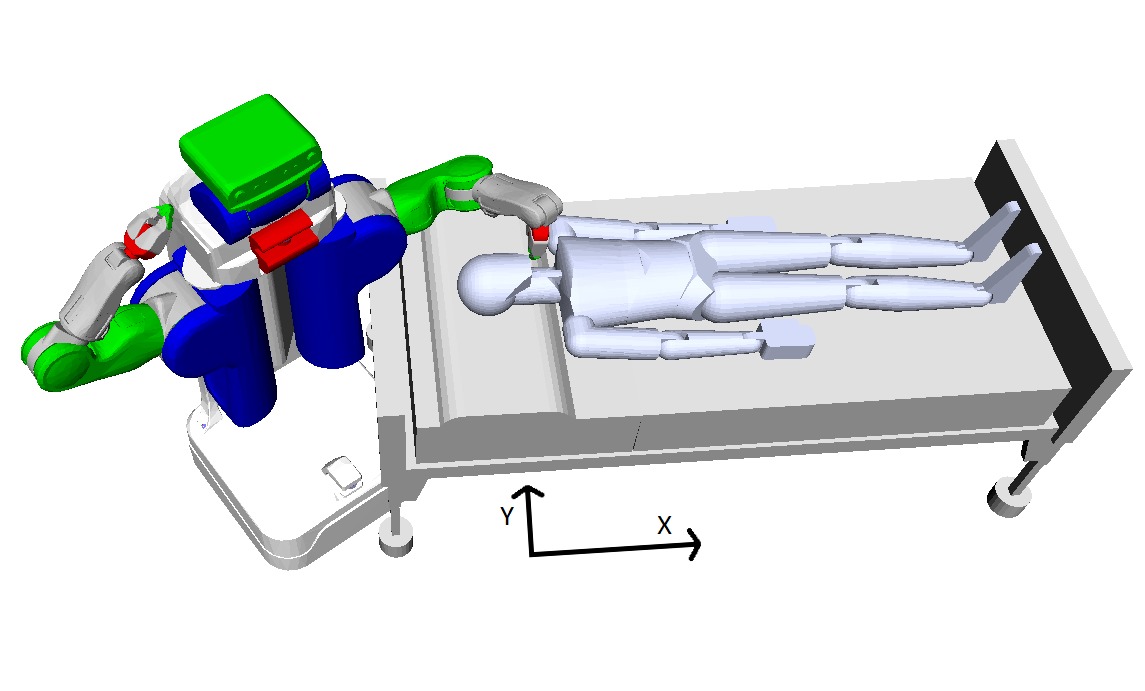}\\
\includegraphics[width=\columnwidth]{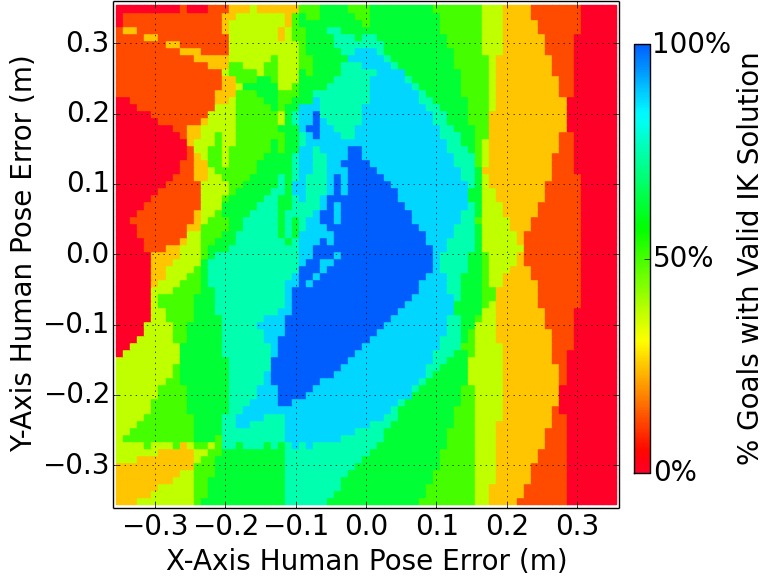}
\caption{The PR2 can perform the shaving task on the person in bed from a single location if there is no wall behind the bed. (Top) the configuration TOC selected (bottom) a visualization of the robustness to error in the person's pose on the bed for this configuration. }
\label{fig:shaving_no_wall}
\end{figure}

TOC may also be used to assist in the design of environments. For example, in our evaluations with 
the robotic bed, TOC selected a configuration for the bed by including the bed's DoF in the robot configuration. 
TOC could similarly optimize many other continuous controllable parameters of the environment.
Comparison of TOC 
scores between a robotic bed and a standard, static bed may demonstrate the value added by a robotic bed. 
By recognizing that for some tasks a lower bed height improves performance and for other tasks a higher bed height improves performance, we may recognize that an adjustable bed may be preferable to allow the bed to reconfigure to the desired height for each task. 
As detailed in Section~\ref{ssec:environment_modeling}, TOC can be used to assist in design for parameters with discrete choices by including it as an 
uncontrollable parameter in the optimization. In this way, you might decide to put the bed against 
the wall, 0.5~m away from the wall, or 1~m away from the wall. For example, we have found that for
the shaving task, if there is sufficient space behind the bed, the PR2 can perform the task from a 
single configuration instead of requiring two configurations. Figure~\ref{fig:shaving_no_wall} 
shows the configuration found by TOC for shaving in the robotic bed without a wall and visualizes the robustness to error in the pose of the person. 

\subsection{Limitations}\label{ssec:limitations}
There are some limitations to our work with TOC and our evaluation.
Although the framework of TOC allows it to search for sets of robot configurations 
of cardinality larger than 2, we limited it to two in our evaluation. We made this choice because we found that more than two robot configurations were not needed for any of the tasks examined. 


We hand designed the task models for our evaluation in simulation, but have not
investigated how well the task models actually represent the tasks. 
In addition, all tasks in this work were
defined with full 6-DoF goal poses, although some tasks, such as sponge baths, have position
requirements and few orientation requirements.
We also did not address tasks with complex motions in which the trajectory between goals poses is important, such as dressing, or tasks with
high strength requirements, such as lifting or ambulating.

Joint-limit-weighted kinematic isotropy, the foundation of our TC-manipulabilty score, does not
account for environmental constraints. 
There may be value in preferring joint configurations away from obstacles. We have mitigated the risk of collisions using a safety margin on the environment and user models at the cost of decreasing the valid 
search space and eliminating valid solutions. Explicitly penalizing proximity to collisions in the objective function, as was 
done by \citet{vahrenkamp2012manipulability}, may be another way to mitigate this issue at the 
cost of additional computation time. 
TOC also does not determine if there are valid paths to reach collision-free IK solutions, which may be problematic when there are environmental constraints. A motion planner could be used to check for valid paths at the cost of computation time.
Although computation time is often a less critical concern for offline processes, it must remain reasonable. We elected to ignore proximity to collisions and the question of the existence of valid paths to achieve lower computation time. 

\section{Conclusion}\label{sec:conclusion}
In this work, we have presented task-centric optimization of robot configurations (TOC), a 
method to select one or more configurations for robots to assist with tasks around a person's body. 
TOC uses TC-reachability and 
TC-manipulability, metrics that we have developed, to represent the robot's dexterity, 
and implicitly handle error. 
TOC is particularly suitable for assistive tasks, where there are a set of
desired tasks known apriori that can be modeled as a set of end effector poses with respect
to relevant reference frames. We have shown that TOC can 
determine a set of one or two robot configurations from which the robot can perform a task well. 
TOC performs the bulk of its computation offline using models of the task, robot, environment,
and person to generate a function that rapidly ($\leq 1$~second) estimates the optimal set of robot configurations
for a task given observations at runtime. We provide evidence that configurations selected by TOC are robust to state estimation errors between the models used offline and
observations at runtime. We created 9 models of assistive tasks to test our system in simulation 
and showed that for each task TOC's average success rate was higher than or comparable to three baseline algorithms from literature. TOC had an overall average success rate of 90.6\% compared to 50.4\%, 43.5\%, and 58.9\% for baseline methods.




\end{document}